\documentclass{article}

\usepackage{microtype}
\usepackage{graphicx}
\usepackage{multirow}
\usepackage{makecell}
\usepackage{tabularx}
\usepackage{multicol}
\usepackage{subcaption}
\usepackage{booktabs} 

\usepackage{hyperref}

\usepackage[preprint]{icml2026}

\usepackage{amsmath}
\usepackage{amssymb}
\usepackage{mathtools}
\usepackage{amsthm}

\usepackage[capitalize,noabbrev]{cleveref}

\theoremstyle{plain}

\theoremstyle{definition}

\theoremstyle{remark}

\usepackage[textsize=tiny]{todonotes}

\icmltitlerunning{Emergent Structured Representations Support In-Context Inference in LLMs}

\begin{document}

\twocolumn[
  \icmltitle{%
  \texorpdfstring{%
    Emergent Structured Representations Support \\
    Flexible In-Context Inference in Large Language Models%
  }{%
    Emergent Structured Representations Support Flexible In-Context Inference in Large Language Models%
  }%
  }
  \icmlsetsymbol{equal}{*}

  \begin{icmlauthorlist}
    \icmlauthor{Ningyu Xu}{yyy}
    \icmlauthor{Qi Zhang}{yyy}
    \icmlauthor{Xipeng Qiu}{equal,yyy}
    \icmlauthor{Xuanjing Huang}{equal,yyy}
  \end{icmlauthorlist}

  \icmlaffiliation{yyy}{College of Computer Science and Artificial Intelligence, Fudan University, Shanghai, China}
  \icmlcorrespondingauthor{Ningyu Xu}{nyxu22@m.fudan.edu.cn}
  \icmlcorrespondingauthor{Xipeng Qiu}{xpqiu@fudan.edu.cn}
  \icmlcorrespondingauthor{Xuanjing Huang}{xjhuang@fudan.edu.cn}

  \icmlkeywords{in-context learning, conceptual representation, reasoning, mechanistic interpretability, large language models}

  \vskip 0.3in
]



\printAffiliationsAndNotice{}  

\begin{abstract}
  Large language models (LLMs) exhibit emergent behaviors suggestive of human-like reasoning. While recent work has identified structured conceptual representations within these models, it remains unclear whether they functionally rely on such representations for reasoning. Here we investigate the internal processing of LLMs during in-context inference across diverse tasks. Our results reveal a conceptual subspace emerging in middle to late layers, whose representational structure persists across contexts. Using causal mediation analyses, we demonstrate that this subspace is not merely an epiphenomenon but is functionally central to model predictions, establishing its causal role in inference. We further identify a layer-wise progression where attention heads in early-to-middle layers integrate contextual cues to construct and refine the subspace, which is subsequently leveraged by later layers to generate predictions. Together, these findings provide evidence that LLMs dynamically construct and use structured latent representations in context for inference, offering insights into the computational processes underlying flexible adaptation.
\end{abstract}

\section{Introduction}

Large language models (LLMs) have developed an emergent capacity for in-context learning, which allows them to adapt to novel tasks specified in context through demonstrations or instructions~\citep{brown_language_2020, lampinen_broader_2025}. This process, often framed as a form of inference, requires no parameter updates~\citep{xie_explanation_2022, hahn_theory_2023, lake_human-like_2023} and arises naturally from the objective of next-token prediction. The in-context learning abilities have sparked immense interest as they showcase traits of human-like learning and generalization~\citep{griffiths_whither_2025, russin_parallel_2025}, enabling impressive, and occasionally human-level, performance across language understanding and reasoning tasks~\citep{wei_emergent_2022, bubeck2023sparks, binz_using_2023, webb_emergent_2023, webb_evidence_2025, lampinen_language_2024, hagendorff_human-like_2023}. However, these successes are tempered by brittleness to contextual variations and non-human-like errors~\citep{lewis_using_2024, dentella_testing_2024, wu_reasoning_2024}. The divergent evidence has fueled ongoing debates over whether LLMs are genuinely approaching structured, human-like inference or merely relying on statistical correlations in their training data~\citep{mitchell_artificial_2025, mitchell_abstraction_2021, mitchell_debate_2023, mccoy_embers_2024, mahowald_dissociating_2024}. Demystifying the internal processes is essential not only for understanding the strengths and limitations of LLM generalization but also for uncovering the computational mechanisms underlying flexible adaptation to changing contexts---a hallmark of natural intelligence~\citep{lake_building_2017}.

Human inference relies on structured knowledge abstracted from experience and adapted to novel situations~\citep{tenenbaum_how_2011, mcclelland_letting_2010, lake_building_2017, johnson-laird_mental_2010}, a capacity requiring conceptual representations that support symbolic processes while accommodating noisy input and graded information~\citep{piantadosi_why_2024}. Recent efforts to probe the internal representations of LLMs have revealed emergent structure that capture core properties of human concepts~\citep{xu_revealing_2025, pavlick_symbols_2023}, but their functional role during inference remains poorly understood. In parallel, while research in mechanistic interpretability has found evidence for symbolic processes supporting in-context learning~\citep{todd_function_2024, yang2025emergent, feucht2025the, yin2025which, olsson2022context}, these findings are primarily based on relatively simple or synthesized tasks and may not generalize across broader contexts~\citep{cho2025revisiting, bakalova2025contextualize, opielka2025analogical}. Consequently, it remains unclear whether LLMs rely on abstract, structured representations for inference and how their internal computations adapt to varying contexts.

\begin{figure*}[t]
    \begin{center}
    \centerline{\includegraphics[width=0.75\textwidth]{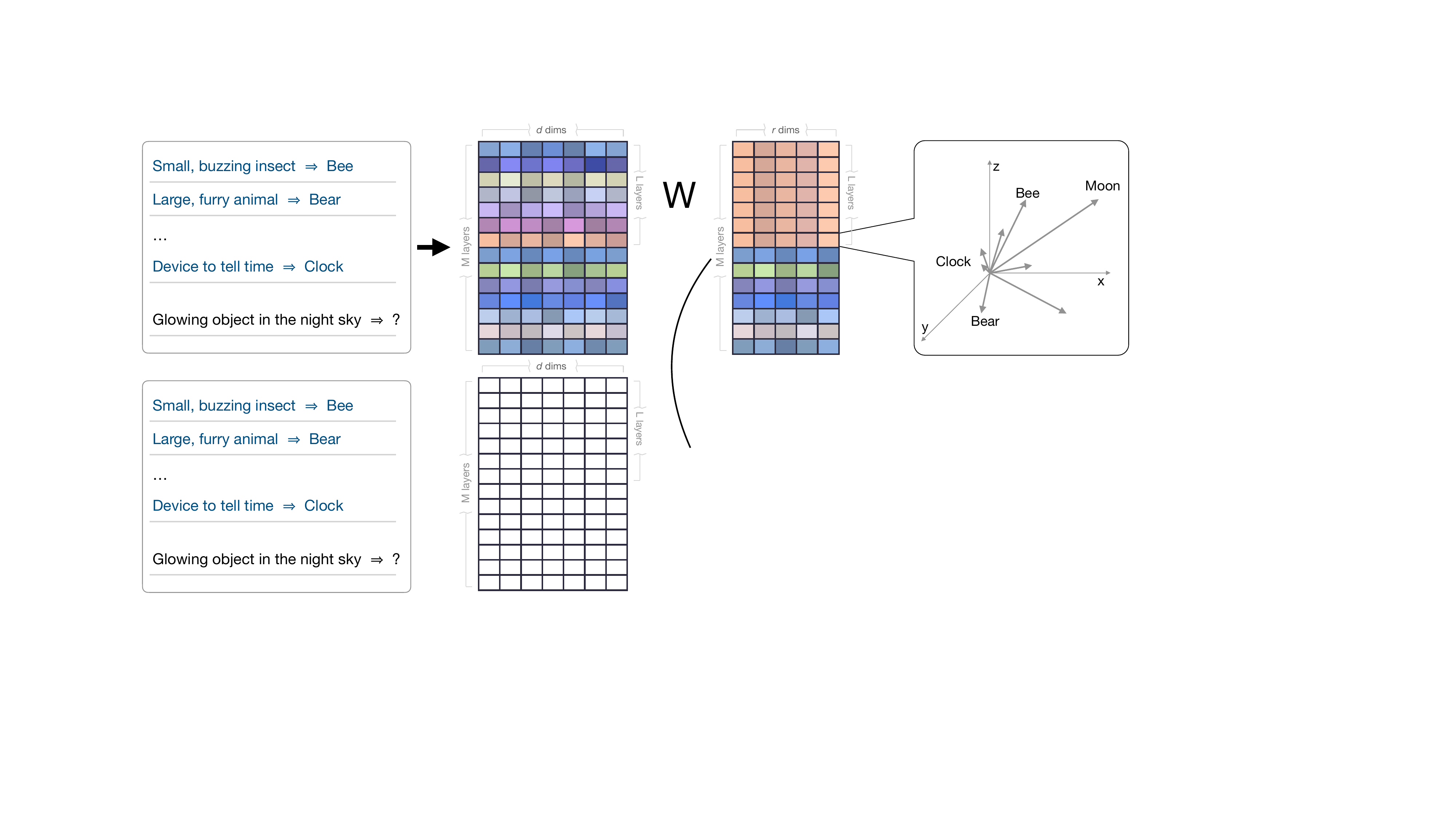}}
    \caption{Illustration of how an emergent conceptual subspace supports in-context inference in LLMs. Given a small set of description--word demonstrations and a query description, a Transformer-based LLM integrates contextual information across layers to form a shared conceptual subspace in the middle-to-late layers. Hidden states can be projected into this subspace, where the relational structure among representations persists across layers and across different demonstration contexts. Causal interventions show that the subspace and its internal relational structure are functionally involved in inference and constrain subsequent computations leading to the final prediction. The concept inference task is shown here as an illustrative example.}
    \label{fig:main}
    \end{center}
\end{figure*}

To address these questions, we applied causal mediation analysis to investigate the internal representations that LLMs construct and use for inference across a diverse set of tasks. We focused on an in-context concept inference task as an informative case study, in which models are guided to derive concepts from descriptions through a few demonstrations (Figure~\ref{fig:main}). This task has been shown to evoke human-like conceptual representations in LLMs, with the resulting representational structure predicting model performance across a range of understanding and reasoning tasks~\citep{xu_revealing_2025}. We therefore used it as a simple but meaningful testbed for probing how structured representations are formed and deployed during reasoning, while also evaluating a broader set of inference tasks drawn from~\citet{todd_function_2024} to assess the generality of our findings.

We first traced when task-relevant information emerges across layers, identifying a conceptual subspace that remains approximately isomorphic from middle to late layers. We then employed targeted causal interventions to explore whether this subspace is causally implicated in the model's inference process. Finally, we examined how this subspace is constructed in context by analyzing how model components---primarily attention heads---write into it. Our experiments reveal an emergent conceptual subspace that encodes structured, task-relevant information and is dynamically recruited during inference. Its internal structure becomes more coherent with additional demonstrations and ultimately persists across contexts. These findings suggest that LLMs do not merely rely on surface-level patterns but construct and deploy abstract, structured representations for in-context inference, offering a mechanistic window into the computational basis of their adaptive behavior.

\section{A Shared Conceptual Subspace Emerges during Inference}

Building on \citet{xu_revealing_2025}, we adopted the reverse dictionary task as a controlled probe to investigate the mechanisms underlying LLMs' in-context inference, while extending our analysis to a broader set of inference tasks from \citet{todd_function_2024} to assess the generality of our findings. The reverse dictionary task simulates the human capacity to identify a specific concept from a description that may be underspecified or vague (Figure~\ref{fig:main}). For example, given an expression such as ``a glowing object in the night sky'' or ``Earth's natural satellite,'' a system must infer the intended concept from context and produce the corresponding term (e.g., ``moon''). Previous work has shown that LLMs can flexibly derive concepts from descriptions in relation to in-context demonstrations, and their penultimate-layer representations exhibit a stable structure that not only predicts downstream model performance but also aligns well with human behavioral and brain data~\citep{xu_revealing_2025}. These properties make this task a well-controlled, informative testbed for investigating whether and how LLMs rely on structured conceptual representations to perform inference in context. However, focusing on the penultimate-layer representations leaves open when and where such structure emerges during internal processing. To address this question, we examined the layer-wise evolution of LLM representations. This analysis reveals a shared conceptual subspace whose internal structure persists from middle to late layers. We then looked into this subspace to understand how representations are organized within it.

\subsection{Methods}

Let $\mathcal{M}$ be an autoregressive Transformer-based LLM with $L$ layers and a hidden dimension of $d$. In the reverse dictionary task, the model is presented with a prompt $s$ consisting of $N$ demonstrations in the format of ``[Description] $\Rightarrow$ [Word]'', followed by a query description and the delimiter ``$\Rightarrow$''. We focused on the computations at the last token position, which is responsible for predicting the target word for the query concept. Let $\mathbf{h}_{\ell}$ denote the hidden state at layer $\ell \in \left\{ 0, \dots, L \right\}$, where $\mathbf{h}_{0}$ corresponds to the initial token embedding. The hidden states evolve through the layers as $\mathbf{h}_{\ell} = \mathbf{h}_{\ell-1} + \sum_{k=1}^{K} \mathbf{a}_{\ell,k} + \mathbf{m}_{\ell}$, where $\mathbf{a}_{\ell,k} \in \mathbb{R}^{d}$ is the output of $k^{\textrm{th}}$ attention head (among $K$ heads) at layer $\ell$, projected to the hidden state~\citep{elhage2021mathematical, todd_function_2024}, and $\mathbf{m}_{\ell}  \in \mathbb{R}^{d}$ is the output of the multilayer perceptron block at layer $\ell$. The final hidden state $\mathbf{h}_{L}$ is mapped by the decoder head to a probability distribution over the model's vocabulary $\mathcal{V}$: $f_{\mathcal{M}} \left(s\right) = \textrm{decode}\left(\mathbf{h}_{L}\right)$, from which the next token is generated.

To characterize the emergence of the representational structure of concepts, we first identified the primary subspace within each layer and compared their relative geometry. For a dataset of $n$ query concepts, let $\mathbf{X}_{\ell} \in \mathbb{R}^{n \times d}$ denote the centered matrix of hidden states at layer $\ell$. We applied singular value decomposition (SVD) to $\mathbf{X}_{\ell}$ and retained the top $k$ principal components that accounted for 95\% of the total variance. We then measured the overlap between layers $\ell$ and $\ell^{\prime}$ by the mean squared cosine of the principal angles $\{\theta_i\}_{i=1}^{k}$ between their respective subspaces: $\frac{1}{k}\sum_{i=1}^{k} \cos^{2}\! \theta_{i}$. This metric captures the extent to which variance-dominant directions are shared across layers, while remaining invariant to basis rotations within each subspace.

Based on the layer-wise alignment patterns, we then employed generalized canonical correlation analysis (GCCA)~\citep{horst_generalized_1961, kettenring_canonical_1971} to isolate a shared subspace that persists across layers. Given hidden-state matrices $\{\mathbf{X}_{\ell}\}_{\ell \in \mathcal{L}}$ from a selected set of layers $\mathcal{L}$, GCCA seeks a common latent representation $\mathbf{G} \in \mathbb{R}^{n \times r}$ that maximizes the alignment across layers. Formally, we solve:
\begin{equation}
    \min_{\mathbf{G}, \{\mathbf{W}_{\ell}\}} \sum_{\ell \in \mathcal{L}} \|\mathbf{G} - \mathbf{X}_{\ell}\mathbf{W}_{\ell} \|_F^2, \quad \text{s.t. } \mathbf{G}^\top\mathbf{G} = \mathbf{I},
\label{eq:gcca}
\end{equation}
where $\mathbf{W}_{\ell} \in \mathbb{R}^{d \times r}$ are layer-specific projection matrices, and $r$ is the dimensionality of the shared subspace. We determined $r$ using a non-parametric permutation test (Appendix~\ref{si:methods-gcca}) to identify a subspace capturing structural alignment beyond noise. Alternative dimensionalities were also evaluated to confirm that the permutation-based choice was appropriate (Appendix~\ref{si:results-subspace-dim}).

Finally, we looked into the internal structure of the identified shared subspace. Beyond GCCA alignment---measured as the average correlation between layer-wise projections onto the shared subspace---we employed two complementary measures to compare subspaces across contexts: (1) subspace overlap, quantified via principal-angle similarity between the projection matrices $\mathbf{W}_{\ell}$, and (2) representational similarity analysis (RSA), which compares the relational structure within each subspace by correlating their (dis)similarity matrices. Detailed definitions and implementation are provided in Appendix~\ref{si:methods-structure}.

We conducted experiments on a range of open-source decoder-only Transformer LLMs spanning multiple model families and scales, including Llama~3.1, Llama~3~\citep{grattafiori2024llama}, and Qwen2.5~\citep{qwen25_2025} (see Appendix~\ref{si:methods-llms} and Table~\ref{tab:llms-details} for details). To probe in-context concept inference, we used materials from the THINGS database~\citep{hebart_things_2019}, which comprises 1,854 concrete, nameable object concepts paired with WordNet synset IDs and definitional descriptions. We also included four tasks from \citet{todd_function_2024}: Antonym, Country--Capital, Landmark--Country, and National Parks (Table~\ref{tab:task-details}). We focus on concept inference in the main text, with results for the remaining tasks reported in Appendix~\ref{si:results-task-generality}.
For each experimental condition, we randomly sampled $N$ description--word pairs as demonstrations, where $N \in \{1, \dots, 48\}$. Results were averaged across five independent runs per condition to ensure robustness (Appendix~\ref{si:methods-data}).

\subsection{Results}
\label{sec:emergence-results}

\begin{figure*}[t]
    \begin{center}
    \centerline{\includegraphics[width=0.9\textwidth]{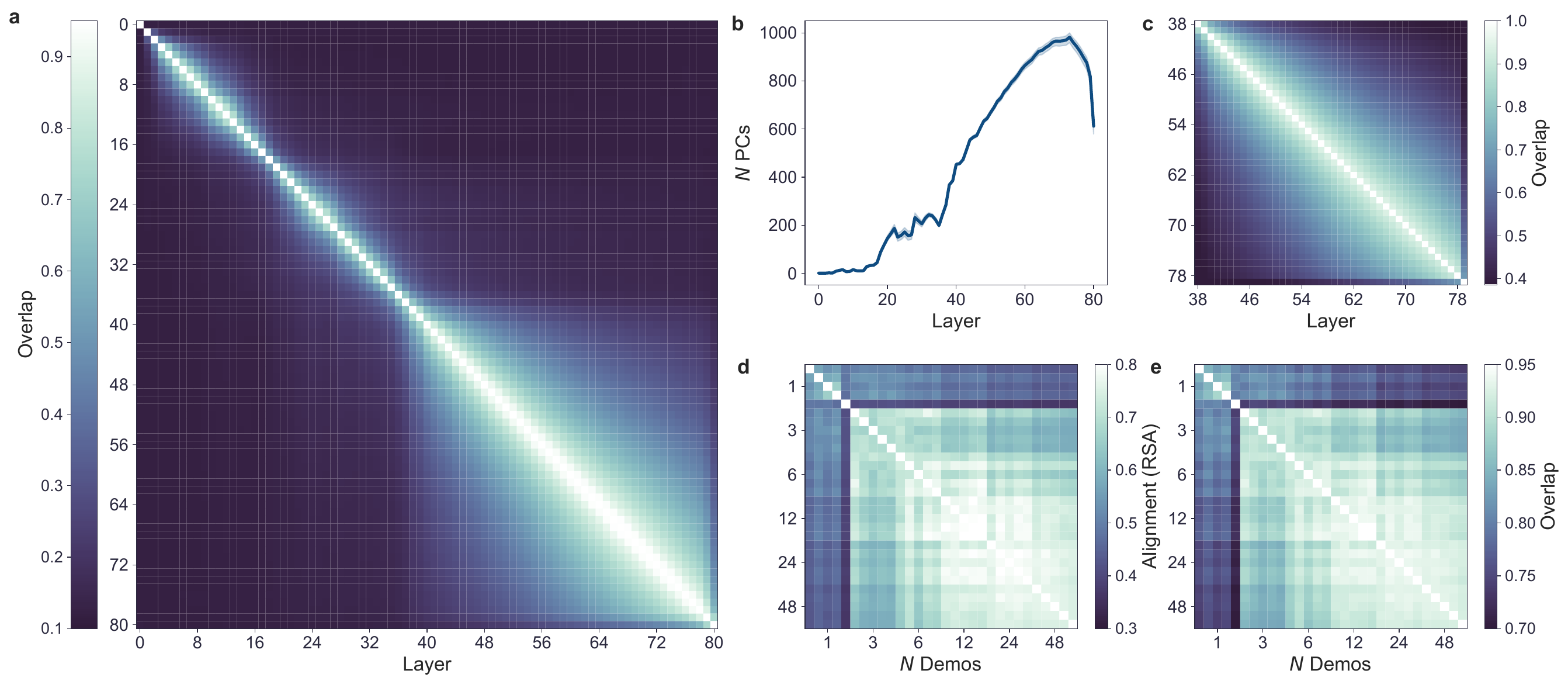}}
    \caption{A shared conceptual subspace emerges in the middle to late layers of Llama-3.1~70B. \textbf{a}, Layer-wise similarity of hidden states, measured as subspace overlap (mean squared cosine of principal angles) between SVD subspaces explaining 95\% variance; averaged over five runs with 24 demonstrations. Axes index layers. \textbf{b}, Number of principal components (PCs) needed to explain 95\% variance across layers, increasing sharply in the middle layers. Shaded areas represent 95\% CIs calculated from 10,000 bootstrap resamples across five runs. \textbf{c}, Overlap between GCCA-derived projection matrices across selected layers. \textbf{d}--\textbf{e}, The conceptual subspace becomes increasingly stable as the number of in-context demonstrations grows. \textbf{d}, Cross-context alignment of representational geometry within the GCCA subspace across demonstration sets, measured by RSA. \textbf{e}, Overlap between GCCA-derived projection matrices across demonstration sets. Axes in \textbf{d}--\textbf{e} indicate the number of demonstrations, with each cell representing a single run.}
    \label{fig:subspace_emerge}
    \end{center}
\end{figure*}

Across all tested models, we observed strong performance on the reverse dictionary task. For Llama-3.1~70B, exact match accuracy increased with the number of demonstrations, rising from 74.56\% ($\pm$4.69\%) with a single demonstration to 87.95\% ($\pm$ 0.32\%) with 24 demonstrations, with marginal gains beyond this point. While smaller models achieved lower accuracy uniformly, this monotonic trend remained consistent across models (Appendix Figure~\ref{fig:si-behavior}), replicating prior findings~\citep{xu_revealing_2025} and confirming that these models can reliably derive target concepts from linguistic descriptions. 

Layer-wise SVD analysis reveals a progressive evolution of models' internal representations across depth (Figure~\ref{fig:subspace_emerge}a). Early layers (up to approximately layers 36--40) exhibited low cross-layer subspace overlap, indicating rapid representational transformation. In contrast, middle to late layers displayed consistently high inter-layer overlap, suggesting the emergence of a shared subspace that is preserved throughout the remainder of the inference process. This transition coincided with a sharp increase in the number of principal components required to explain 95\% of the variance (Figure~\ref{fig:subspace_emerge}b), suggesting that contextual information is integrated into a higher-dimensional subspace for subsequent computation. Together, these results indicate a qualitative shift in models' internal processing, where most representational structure is established by mid-depth, after which subsequent layers primarily operate within this subspace rather than introducing new dominant directions.

Based on this observed stability, we applied GCCA to isolate a shared subspace spanning the middle to late layers. Both GCCA alignment scores and RSA values approached unity across these layers, indicating strong linear correspondence. At the same time, the layer-specific projection matrices exhibited gradual variation, with higher similarity between adjacent layers (Figure~\ref{fig:subspace_emerge}c). This pattern suggests that later layers preserve a common representational structure while iteratively refining the subspace to generate predictions.

Finally, we examined how this shared subspace varies across different contexts. As the number of demonstrations increases from 1 to 24, alignment between the internal structure of the subspaces increased steadily, with diminishing improvements beyond this threshold (Figure~\ref{fig:subspace_emerge}d). A similar trend was observed in the overlap between subspaces extracted from the same layer across contexts (Figure~\ref{fig:subspace_emerge}e). While this pattern held across models, larger models such as Llama-3.1~70B consistently exhibited higher cross-context alignment (Appendix~\ref{si:results-emergence}). These findings suggest that, as LLMs receives more contextual cues, they construct an increasingly coherent and context-invariant relational structure during inference. Such invariance indicates abstraction beyond surface-level details and provides a representational basis for generalization across contexts. 

\section{The Conceptual Subspace Causally Mediates Inference}

We next examined the functional role of the emergent conceptual subspace in model inference through a series of causal interventions. First, we investigated the contextual information mediated by the subspace through causal mediation analysis under targeted input corruptions~\citep{pearl2001direct, meng_locating_2022, wang2023interpretability, geva2023dissecting, todd_function_2024}. We then assessed the necessity and sufficiency of the subspace for inference through selective ablation and isolation. Finally, we assessed whether the relational structure encoded in the subspace generalizes across contexts, enabling subspace-level information from one context to modulate inference in another.

\subsection{Methods}
\label{sec:causal-methods}

We define an orthogonal projection operator for each layer $\ell$ as $\mathbf{P}_{\ell} = \mathbf{W}_{\ell}\mathbf{W}_{\ell}^{\top}$, where $\mathbf{W}_{\ell} \in \mathbb{R}^{d \times r}$ is the orthonormal basis spanning the GCCA-derived subspace. This operator allows us to decompose any hidden state $\mathbf{h}_{\ell} \in \mathbb{R}^d$ into the subspace component $\mathbf{h}_{\ell,\parallel} = \mathbf{P}_{\ell}\mathbf{h}_{\ell}$ and its orthogonal complement, $\mathbf{h}_{\ell,\perp} = (\mathbf{I} - \mathbf{P})\mathbf{h}_{\ell}$. We employed this decomposition throughout the intervention analyses. To contextualize the causal effect of the subspace, we compared it with randomly generated orthonormal bases of the same dimensionality.

We performed activation patching to characterize the contextual information mediated by the conceptual subspace. This intervention estimates the causal contribution of subspace-level representations by transplanting activations from a clean context into a corrupted one. Let $s = \left[\left(x_{1}, y_{1}\right), \dots, \left(x_{N}, y_{N}\right), x_{q}\right]$ denote a prompt with $N$ demonstrations, where $x_{i}$ and $y_{i}$ ($i = 1, \dots, N$) are the description and word in the demonstrations, and $x_{q}$ is the query description, with $y_{q}$ denoting its expected correct term. We considered three corruption conditions $\tilde{s}$ to probe different information streams: (i) description corruption ($\tilde{x}_i$), (ii) label (word) corruption ($\tilde{y}_i$), and (iii) query corruption ($\tilde{x}_q$) (Figure~\ref{fig:si-corruption}).

During a forward pass on $\tilde{s}$, we replaced only the subspace-aligned component of the corrupted state $\mathbf{h}_{\ell}^{\textrm{corr}}$ with that of the clean state $\mathbf{h}_{\ell}^{\textrm{clean}}$: $\mathbf{h}_{\ell}^{\textrm{patch}} = \mathbf{h}_{\ell}^{\textrm{corr}} + \mathbf{P}_{\ell}(\mathbf{h}_{\ell}^{\textrm{clean}} - \mathbf{h}_{\ell}^{\textrm{corr}})$.
We evaluated the effect of this intervention using the causal indirect effect \citep[CIE;][]{todd_function_2024}, calculated as
\begin{equation}
    \mathrm{CIE}\!\left(\mathbf{h}_{\ell,\parallel}\right) = \log f_{\mathcal{M}}\!\left(\tilde{s} \mid \mathbf{h}_{\ell}^{\textrm{patch}} \right)\!\left[y_{q}\right] - \log f_{\mathcal{M}}\!\left(\tilde{s} \right)\!\left[y_{q}\right],
\end{equation}
where $f_{\mathcal{M}}\!\left(\cdot\right)\!\left[y_{q}\right]$ denotes the output probability for token $y_{q}$ from model $\mathcal{M}$. A positive CIE indicates that restoring the subspace component recovers task-relevant information disrupted by the corruption. We also computed the normalized CIE to quantify the fraction of performance recovered:
\begin{equation}
    \mathrm{NormCIE}\!\left(\mathbf{h}_{\ell,\parallel}\right) = \frac{\mathrm{CIE}\!\left(\mathbf{h}_{\ell,\parallel}\right)}{\log f_{\mathcal{M}}\!\left(s\right)\!\left[y_{q}\right] - \log f_{\mathcal{M}}\!\left(\tilde{s}\right)\!\left[y_{q}\right]}.
\end{equation}
This allows us to dissociate the types of contextual information routed through the subspace and to localize their effects across layers.

To assess whether the conceptual subspace is necessary for inference, we performed subspace ablation by removing the component of the hidden state lying in the subspace, while leaving the orthogonal complement unchanged: $\mathbf{h}_{\ell} \leftarrow \left(\mathbf{I} - \mathbf{P}_{\ell}\right) \mathbf{h}_{\ell}$. Similarly, to test for sufficiency, we performed subspace isolation by retaining only the subspace component: $\mathbf{h}_{\ell} \leftarrow \mathbf{P}_{\ell}\mathbf{h}_{\ell}$. The effect of these interventions was measured by the change in the log-probability of the correct term: $\log f_{\mathcal{M}}\!\left(s\mid \mathbf{h}_{\ell}^{\prime}\right)\!\left[y_{q}\right] - \log f_{\mathcal{M}}\!\left(s\right)\!\left[y_{q}\right]$, where $\mathbf{h}_{\ell}^{\prime}$ denotes the modified state. A substantial performance drop after ablation indicates that the subspace is causally necessary for inference, while high performance preservation after isolation suggests that the subspace alone contains sufficient information to support inference.

Finally, to test whether inference relies on a context-invariant relational structure encoded in the conceptual subspace, we performed cross-context subspace transfer. For a given pair of source and target contexts $c_{i}$ and $c_{j}$, each defined by a set of $N$ demonstrations, we learned an orthogonal transformation $\mathbf{Q}_{\ell}^{\left(c_i,c_j\right)} \in \mathbb{R}^{r \times r}$ between their subspaces using a training set of query concepts. We then evaluated cross-context transfer on held-out query concepts. For a shift from concept $q_b$ to $q_a$, we computed the vector offset in the source context $c_i$: $\Delta_{\ell, \left(q_{a},q_{b}\right)}^{c_{i}} = \mathbf{W}_{\ell}^{c_{i} \top}\left(\mathbf{h}_{\ell,q_{a}}^{c_{i}} - \mathbf{h}_{\ell,q_{b}}^{c_{i}}\right)$. This offset was then mapped to the target context $c_{j}$ for query concept $q_{b}$: $\mathbf{h}_{\ell,q_{b}}^{c_{j}} \leftarrow \mathbf{h}_{\ell,q_{b}}^{c_{j}} + \mathbf{W}_{\ell}^{c_{j}} \mathbf{Q}_{\ell}^{\left(c_i,c_j\right)} \Delta_{\ell, \left(q_{a},q_{b}\right)}^{c_i}$. We quantified the effect of this intervention using the causal mediation score~\citep{wang2023interpretability}, which measures the change in the log probability difference between $y_{q_a}$ and $y_{q_b}$ after patching (Appendix~\ref{si:methods-cma}). Successful modulation under this transfer indicates that the conceptual subspace supports inference through a functionally aligned relational structure that generalizes across contexts. Besides random orthonormal bases, we additionally compared cross-context transfer against patching using the conceptual subspace derived from the same context, which serves as a within-context reference condition.

\begin{figure*}[t]
    \begin{center}
    \centerline{\includegraphics[width=0.9\textwidth]{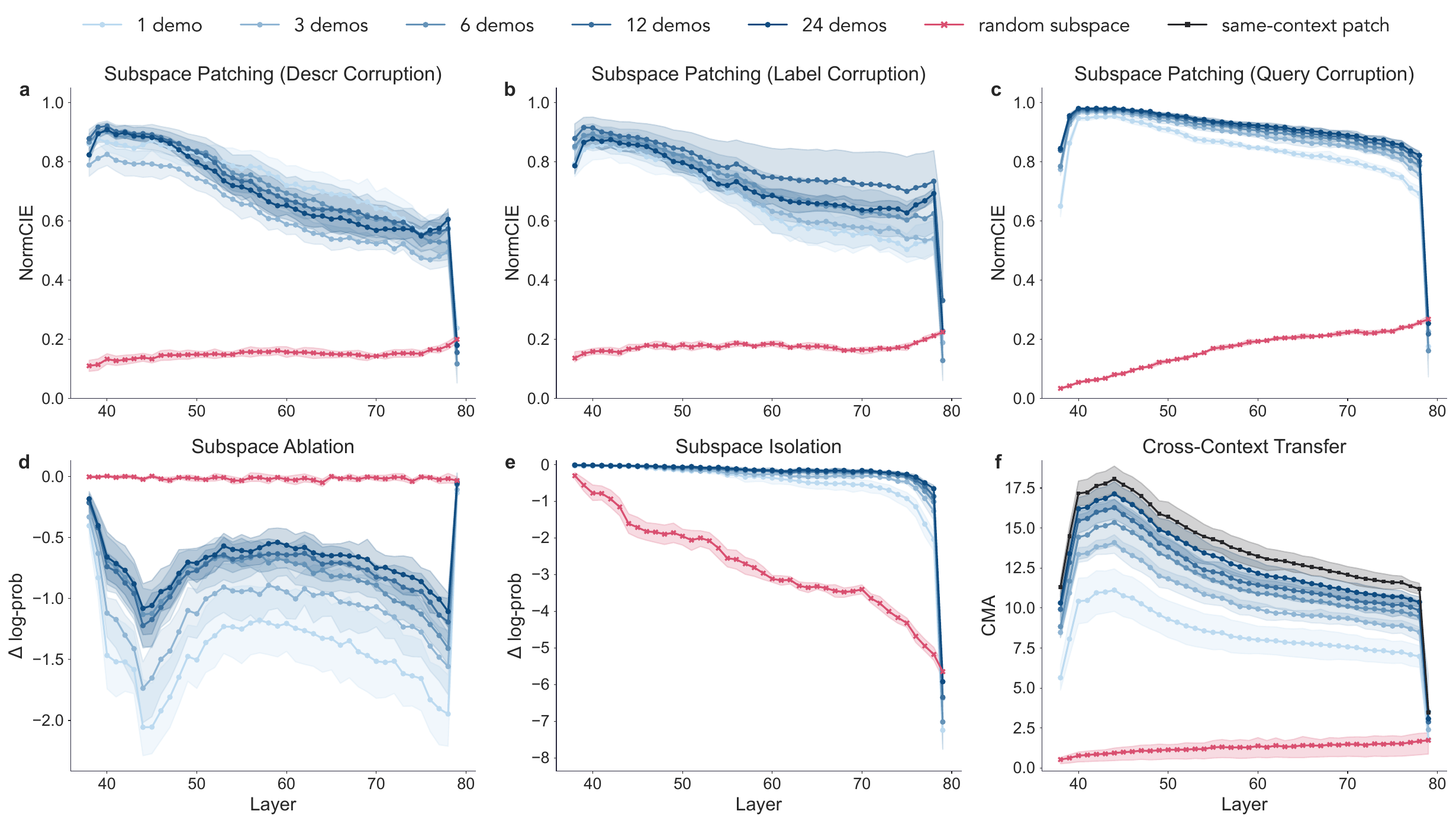}}
    \caption{The conceptual subspace causally mediates model inference. \textbf{a}--\textbf{c}, Activation patching with $N$ demonstrations under three corruption conditions: description (\textbf{a}), label (\textbf{b}) and query (\textbf{c}). Patching the conceptual subspace (blue) is compared against a random-subspace baseline (red). The x-axis indexes layers, and the y-axis shows the normalized causal indirect effect (CIE). \textbf{d}--\textbf{e}, Subspace necessity and sufficiency tested by ablating the conceptual subspace (\textbf{d}) or isolating it (\textbf{e}). The y-axis denotes the change in log-probability of the correct token. \textbf{f}, Causal effects of cross-context transfer, where the relational structure from a source context is adapted to a target context via an orthogonal transformation. The y-axis reports the causal mediation (CMA) score. Performance is compared against a random-subspace baseline (red) and a same-context patch reference (black) using the conceptual subspace derived from the target context itself. Shaded regions indicate 95\% CIs computed from 10,000 bootstrap resamples across five runs.}
    \label{fig:causality}
    \end{center}
\end{figure*}

\subsection{Results}
\label{sec:causal-results}

Figure~\ref{fig:causality} summarizes the effects of targeted causal interventions in Llama-3.1~70B. Activation patching under corrupted demonstrations (Figure~\ref{fig:causality}a--b) shows that model performance can be largely restored by patching the conceptual subspace into middle layers. With 24 in-context demonstrations, the normalized CIE peaked around layer 40 for both description corruption (90.92\% $\pm$ 1.28\%) and label corruption (87.81\% $\pm$ 2.43\%), substantially surpassing the random-subspace baseline. Recovery then decreased gradually in later layers, with a sharp drop at layer 79. These results indicate that demonstration-derived contextual information is primarily mediated by the conceptual subspace at mid-depth, where it is integrated for subsequent computation. Patching later layers instead engages computations driven by the corrupted context, which do not fully rely on the conceptual subspace and yield reduced recovery.

Under query corruption (Figure~\ref{fig:causality}c), recovery was even more pronounced and sustained, with normalized CIE peaking at 98.15\% ($\pm$0.46\%) around layer 42 and remaining robust through layer 78 (82.24\% $\pm$ 1.72\%). Compared to demonstration corruption, the effect spanned a broader and later window, suggesting that query processing relies on the same conceptual subspace but persists longer in the residual stream. The final-layer decline across all conditions likely reflects a transition from abstract processing to surface-form generation, where the conceptual subspace is less relevant.

Subspace necessity and sufficiency were assessed via ablation and isolation (Figure~\ref{fig:causality}d--e). Ablating the conceptual subspace produced a marked performance drop relative to the random baseline, with the largest effects around layers 44--45 and 77--78 (changes in log-probability exceeding 1), showing that removing this information disrupts inference. However, performance did not collapse entirely, indicating partial redundancy or compensatory pathways. Conversely, isolating the subspace was sufficient to maintain near-original performance across most layers, with degradation confined to the final layers. As the number of demonstrations increases, ablation effects became less pronounced, while isolation continued to preserve performance deeper in the model, suggesting a more robust reliance on the subspace as contextual information accumulates.

Finally, for cross-context transfer, the causal mediation (CMA) score peaked around layer 44 (17.13 $\pm$ 0.90) and remained significant through layer 78 (10.36 $\pm$ 0.38), mirroring the layer-wise profile observed for query corruption. The effect increased with the number of demonstrations, closely approaching the same-context patch reference while substantially exceeding the random-subspace baseline. Together, these results indicate that the model progressively constructs and exploits a more stable, abstract conceptual subspace as contextual cues accumulate, which supports the generalization of relational structure across contexts. Comparable effects were observed across a broader set of inference tasks (Appendix~\ref{si:results-task-generality}) and were noticeably more pronounced in larger models such as Llama-3.1~70B (Appendix~\ref{si:results-causality}), suggesting that increased scale and compute facilitate the emergence and utilization of such abstract representations for approximating in-context inference.

\section{Contextual Information Shapes the Conceptual Subspace}

How is the emergent conceptual subspace constructed based on contextual information? To address this question, we focused on attention heads, which are the primary mechanism by which Transformer models integrate contextual information~\citep{vaswani_attention_2017, elhage2021mathematical}. We employed causal mediation analysis to isolate attention heads critical in inference, and then characterized their role by analyzing their attention patterns and output representations, probing how they contribute to constructing and refining the conceptual subspace. 

\begin{figure*}[t]
    \begin{center}
    \centerline{\includegraphics[width=\textwidth]{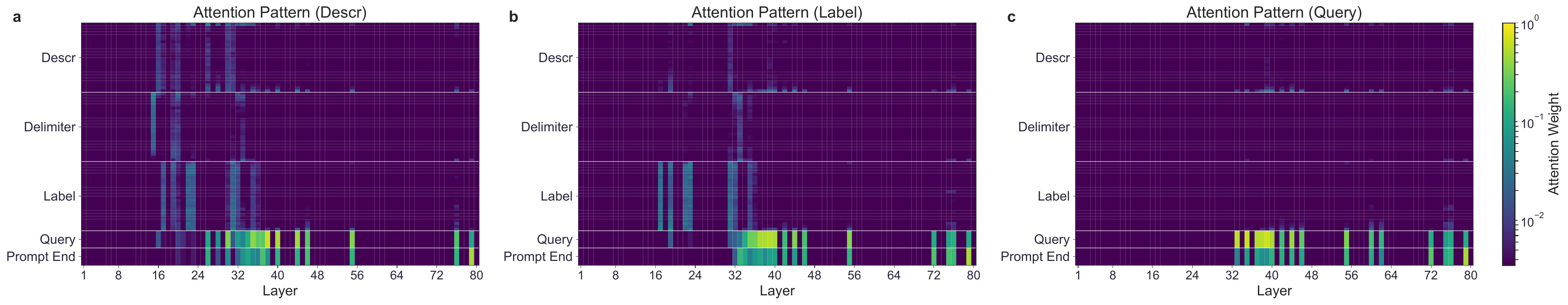}}
    \caption{Attention patterns of heads with statistically significant causal indirect effects (CIEs) identified under description (\textbf{a}), label (\textbf{b}), and query corruption (\textbf{c}). Within each layer, attention patterns were averaged over heads with statistically significant CIEs. Layers with no significant heads were assigned zero values. The x-axis indicates layer index, and the y-axis shows attended token spans grouped by source segment. The description, delimiter, and label spans for the 24 in-context demonstrations are ordered top-to-bottom based on their sequence in the prompt. See Figure~\ref{fig:si-attn_pattern}a--c for attention patterns of individual heads.}
    \label{fig:attn_pattern}
    \end{center}
\end{figure*}

\begin{figure*}[t]
    \begin{center}
    \centerline{\includegraphics[width=0.9\textwidth]{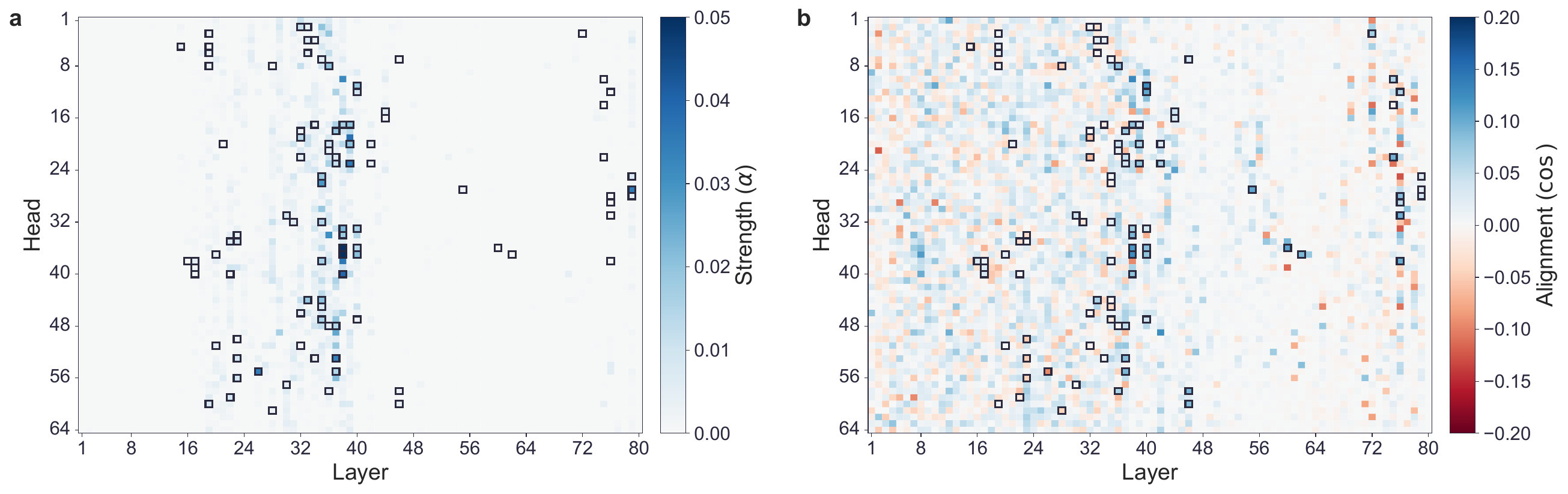}}
    \caption{Contribution of attention heads to the conceptual subspace in Llama-3.1~70B. \textbf{a}, Contribution strength ($\alpha$) of attention heads to the conceptual subspace. \textbf{b}, Directional alignment (cosine similarity) between attention head outputs and the conceptual subspace. Bordered cells highlight attention heads with statistically significant causal indirect effects (CIEs).}
    \label{fig:attn_alpha_align}
    \end{center}
\end{figure*}

\subsection{Methods}

We used activation patching to identify the attention heads that are functionally relevant for inference~\citep{todd_function_2024, yang2025emergent}. Following the procedure in Section~\ref{sec:causal-methods}, we constructed three corruption conditions $\tilde{s}$ including description, label, and query corruption. Let $\mathbf{a}_{\ell,k} \in \mathbb{R}^{d}$ represent the output of the $k^{\textrm{th}}$ attention head at layer $\ell$, projected into the residual stream. For each corrupted input, we intervened by replacing the corrupted head output $\mathbf{a}_{\ell,k}^{\text{corr}}$ with its counterpart from the clean run $\mathbf{a}_{\ell,k}^{\text{clean}}$. The causal effect of this intervention was then measured via the causal indirect effect (CIE). We performed a permutation test to control the family-wise error rate and retained only heads with statistically significant effects ($p < 0.05$; Appendix~\ref{si:methods-cma}).

We then probed the attention patterns of the identified heads to characterize what contextual information they integrate. For each head, we aggregated attention mass across five token spans: (i) demonstration descriptions, (ii) mapping delimiters (e.g., ``$\Rightarrow$''), (iii) demonstration labels, (iv) query tokens, and (v) the final prompt delimiter preceding the prediction. This allows us to localize the specific contextual information the head preferentially attends to during inference.

Finally, to assess the contribution of the identified heads to the shared conceptual subspace, we analyzed both the magnitude and directional consistency of their updates. Let $\mathbf{W}_{\ell} \in \mathbb{R}^{d \times r}$ denote the GCCA-derived projection matrix for layer $\ell$, and $\mathbf{Y}_{\ell} = \mathbf{X}_{\ell}\mathbf{W}_{\ell} \in \mathbb{R}^{n \times r}$ be the projected hidden states of $n$ concepts. For a head $k$ at layer $\ell$ with stacked outputs $\mathbf{A}_{\ell,k} \in \mathbb{R}^{n \times d}$, we define the head-induced subspace update as: $\Delta \mathbf{Y}_{\ell, k} := \mathbf{A}_{\ell,k} \mathbf{W}_{\ell} \in \mathbb{R}^{n \times r}$. To characterize the nature of these updates, we computed two complementary metrics: (i) contribution strength ($\alpha_{\ell,k}$) and (ii) directional alignment ($\text{align}_{\ell,k}$). Contribution strength measures the magnitude of the head's update relative to the total subspace energy: $\alpha_{\ell,k} = \frac{\|\Delta\!\mathbf{Y}_{\ell, k}\|_{F}^{2}}{\|\mathbf{Y}_{\ell, k}\|_{F}^{2}}$, and directional alignment quantifies the geometric coherence between the head's update and the target subspace representation by their cosine similarity: $\textrm{align}_{\ell,k} = \frac{\left \langle \Delta\!\mathbf{Y}_{\ell, k}, \mathbf{Y}_{\ell, k}  \right \rangle}{\|\Delta\!\mathbf{Y}_{\ell, k}\|_{F} \|\mathbf{Y}_{\ell, k}\|_{F}}$, where $\left \langle \cdot, \cdot \right \rangle$ denotes the Frobenius inner product. For heads in layers $\ell$ preceding the first available subspace basis at layer $\ell^{\ast}$, we calculated $\alpha_{\ell,k}$ and $\textrm{align}_{\ell,k}$ relative to $\mathbf{W}_{\ell^{\ast}}$. This approach was motivated by the hypothesis that information written to the residual stream can persist and remain linearly accessible to subsequent layers.

\subsection{Results}

Activation patching across corruption conditions reveals a progression of information integration (Figures~\ref{fig:attn_pattern}--\ref{fig:attn_alpha_align}; Appendix Figure~\ref{fig:si-sgnf_heads-llama3}a--c). Under demonstration corruptions (descriptions and labels), statistically significant heads were concentrated in the early-to-middle layers (17--40), coinciding with the depth at which the shared conceptual subspace emerged and began to causally mediate contextual information (around layers 36--40; Section~\ref{sec:causal-results}). The substantial overlap between heads identified under description and label corruptions suggests a joint mechanism that integrates these components to construct an abstract conceptual subspace, with description corruption yielding stronger CIEs (Appendix Figure~\ref{fig:si-sgnf_heads-llama3}a--c), likely due to its role in contextualizing the query~\citep{bakalova2025contextualize}. In contrast, query corruption implicated a later set of heads spanning the middle to late layers, mirroring our observation that query information is mediated by the conceptual subspace over an extended stage of processing. Mid-layer heads (approximately 33--42) likely support query-related computations within the subspace, whereas the final-layer heads (79--80) may implement a readout stage that converts subspace representations into surface-form token predictions.

Attention mass attribution reveals a functional shift from demonstration-centric to query-centric processing (Figure~\ref{fig:attn_pattern}). Heads identified under demonstration corruptions and located before layer 38 allocated most attention to contextual descriptions, delimiters, and labels, consistent with a role in extracting and integrating demonstration-related cues to construct the subspace. Beyond roughly layer 40, these heads shifted attention toward the query and the final prompt delimiter, signaling a transition from subspace construction to query-conditioned inference. Heads identified under query corruption exhibited heavy attention to query tokens, while the final-layer heads attended predominantly to the last delimiter to facilitate next-token prediction.

The identified heads were also geometrically aligned with the conceptual subspace (Figure~\ref{fig:attn_alpha_align}). Significant heads in early-to-middle layers exhibited both higher contribution strength ($\alpha$) and greater directional alignment ($\textrm{align}$) compared to non-significant heads. While contribution strength decreased in later layers, directional alignment remained high, suggesting that late-layer computations stay coordinated with the already-established subspace. Together, these results suggest a two-stage process: (i) Construction, where early-to-middle layers integrate demonstration-derived cues to build and stabilize the conceptual subspace; and (ii) Utilization, where subsequent layers operate within this structured subspace to support inference and produce the final prediction. Results from other LLMs support the generality of our findings (Appendix~\ref{si:results-context}).

\section{Related Work}

The in-context learning capacity of LLMs~\citep{brown_language_2020, lampinen_broader_2025} has been studied extensively from both theoretical and empirical perspectives. Theoretically, in-context learning has been framed as approximate Bayesian inference~\citep{xie_explanation_2022, hahn_theory_2023, wies2023learnability} or as a form of meta-learning emerging from next-token prediction~\citep{akyurek2023what, chan_data_2022, mahankali2024one}. Empirically, research has investigated how various contextual factors can shape model behavior and generalization~\citep{min2022rethinking}. Hybrid in-context and in-weight learning in neural networks has been shown to capture aspects of human learning and generalization \citep{russin_parallel_2025, lake_human-like_2023}. While these lines of work characterize what problems in-context learning can solve, the internal machinery that implement these computations remains a subject of active debate.

Mechanistic interpretability seeks to decompose LLM computation into interpretable components and circuits~\citep{elhage2021mathematical, meng_locating_2022, bakalova2025contextualize, cho2025revisiting}. A growing body of work has identified mechanisms supporting abstract or symbolic-like operations during in-context learning~\citep{olsson2022context, hendel2023context, todd_function_2024, feng2024how, feucht2025the, yang2025emergent}. However, much of this evidence comes from relatively simple, synthetic, or tightly controlled settings. Our work complements this literature by moving beyond individual attention heads to characterize the representational substrate of inference. Specifically, we identify an abstract conceptual subspace formed in the residual stream that captures structured, task-relevant information, generalizes across contexts, and is causally implicated in inference over real-world knowledge.

Our work also builds on previous work probing the internal representations learned by LLMs. Evidence suggests that these models encode substantial conceptual knowledge~\citep{forbes2019neural, xu_revealing_2025} and can dynamically form representations of world state~\citep{gurnee2023language, li2023emergent, park2025iclr}. Despite this progress, it remains unclear whether and how such structured representations are formed and used during inference. We bridge this gap by linking representational structure to causal mechanisms, demonstrating how LLMs construct and leverage a conceptual subspace to perform in-context inference.

\section{Discussion}

In this work, we identify an emergent representational structure that bridges raw contextual input and abstract inference. Our results demonstrate that LLMs construct a dedicated conceptual subspace that abstracts away surface-level details while organizing concepts into a coherent relational structure. This structure persists across contexts and is functionally engaged during inference. These findings inform ongoing debates about the capacity of LLMs for genuine inference: despite being trained solely for next-token prediction, they can learn internal representational substrates that approximate abstract, structured inference rather than relying exclusively on superficial heuristics.

Prior work suggests that middle layers contain substantial structured knowledge and encode meaningful aspects of world state~\citep{skean2025layer, park2025iclr}. Our results extend this view by showing how contextual information and stored knowledge are integrated across middle to late layers to form a coherent conceptual structure that later computations can exploit for task-specific inference. A natural next step is to trace how information encoded in the conceptual subspace is transformed into the target output, and to determine whether analogous substrates support broader forms of reasoning beyond the inference tasks studied here. More generally, connecting such mechanisms to theories of concepts can help clarify how structured inference and reasoning can emerge in both humans and machines~\citep{piantadosi_why_2024, johnson-laird_mental_2010}.

While the conceptual subspace is sufficient and causally central to inference, ablation results also reveal redundancy and compensatory pathways. This raises the possibility that models implement partially overlapping routes to the same behavioral outcome, an important consideration for mechanistic claims and for connecting distributed computation to the symbolic or structured procedures it can approximate. Moreover, our results on smaller models suggest that clean reliance on an abstract conceptual subspace is emergent and scale-dependent. Future work could characterize when models learn to construct and use such an abstract subspace versus alternative pathways, and whether these routes differ in robustness, sample efficiency, or generalization profile.

Our findings also highlight that the mechanisms supporting inference---and the functional roles of model components---shift systematically as contextual evidence accumulates~\citep{lampinen2026linear, lubana2026priors}. Understanding how internal processing changes across contexts, and how post-training interventions alter models' dependence on contextual cues, is therefore an important direction both for theory-building and for predicting how models will generalize across scenarios, with direct implications for AI safety and reliability.

Finally, our work provides a generalizable framework for studying the internal computation of LLMs at the level of representations, complementing circuit-level analyses that focus on individual components such as heads or neurons. By combining subspace identification with causal intervention, this approach can be naturally extended to other tasks and models. The resulting functionally meaningful subspaces further open a promising interface for steering and monitoring. In particular, our cross-context subspace transfer experiment provides a proof of concept for interventions at the level of task-relevant representations. Exploring whether and how such subspaces can enable flexible, high-level steering of AI systems in more realistic applications and on more complex tasks is a compelling direction for future research~\citep{beaglehole2026toward}.

\paragraph{Limitations}

Our work uses concept inference as a simple but informative testbed for probing mechanisms underlying flexible inference in LLMs, which has been shown relevant to broader language understanding and reasoning abilities~\citep{xu_revealing_2025, xu2024tip}. Although our results suggest that the identified mechanisms generalize beyond this setting to a broader family of inference tasks, we do not yet establish their scope in more complex, real-world tasks, which will be essential for building a fuller picture of model competence and limitations.

We used SVD followed by GCCA as a conservative operational procedure for identifying shared cross-layer representational structure, whose cross-context stability and functional relevance were then validated through representational and causal analyses. While this choice provides an interpretable estimate of shared structure, it inherently favors linear, high-variance, and cross-layer-consistent representations, potentially overlooking task-relevant information that is nonlinear, low-variance, layer-specific, or more distributed. Although sufficient for our claims, future work could explore complementary methods for isolating relevant subspaces, such as nonlinear approaches, task-conditioned subspace discovery, and techniques for capturing multiple interacting latent factors. Our results further suggest that, with sufficient in-context demonstrations, relevant subspaces become increasingly aligned, motivating future work on extracting cross-context shared subspaces for analysis and steering.


\section*{Impact Statement}

This work advances scientific understanding of how large language models perform flexible in-context inference. By showing that models organize task-relevant information into a latent conceptual subspace whose relational structure persists across contexts and is causally involved in inference, it may inform future methods for interpretability, monitoring, and intervention. In turn, such methods may improve analysis of model reasoning and failure modes, and support the development of more reliable and controllable AI systems.

At the same time, techniques for identifying and manipulating internal representations may present dual-use risks. More precise intervention in model behavior could support beneficial applications, including safety and alignment, but could also be misused. Our work is primarily foundational rather than a deployable intervention method, though it may help enable more fine-grained analysis and control of model computation.

We expect the main societal value of this work to lie in improving understanding of large language models and supporting future research on transparency, safety, and reliability.


\bibliography{reference}
\bibliographystyle{icml2026}

\newpage
\appendix
\onecolumn

\section{Data, Code, and Materials Availability}

This work uses publicly available datasets and materials~\citep{hebart_things_2019, todd_function_2024, nguyen_distinguishing_2017, hernandez2024linearity}. Code for reproducing all experiments is available at \url{https://github.com/ningyuxu/llm_structured_icl}.

\section{Supplementary Methods}

\subsection{Models}
\label{si:methods-llms}

Our experiments focus on base models---LLMs pretrained solely with next-token prediction, without any post-training or instruction tuning. Table~\ref{tab:llms-details} summarizes the open-source LLMs included in our study. In the main paper, we report results for Llama-3.1~70B, which is the largest model we evaluate and the one that achieves the strongest behavioral performance on the reverse dictionary task, making it a natural representative for illustrating the emergence of the conceptual subspace. Results for the remaining models are provided in Appendix~\ref{si:results-emergence}--\ref{si:results-context} and Figures~\ref{fig:si-subspace_emerge-llama3}--\ref{fig:si-attn_alpha_vs_align}.

\subsection{Data}
\label{si:methods-data}

For concept inference, we used data from the THINGS database~\citep{hebart_things_2019} (available at \url{https://osf.io/jum2f/}), which contains 1,854 concrete, nameable object concepts selected to be representative of everyday objects commonly used in American English. Each concept is paired with its WordNet synset ID, definitional descriptions, and multiple associated images, providing a suitable testbed for analyzing conceptual representations derived from natural-language descriptions.

We further extended our analyses to four additional inference tasks from \citet{todd_function_2024}: Antonym, Country--Capital, Landmark--Country, and National Parks (Table~\ref{tab:task-details}).

To probe in-context concept inference in LLMs, we randomly designated 20\% of the concepts as a training set, from which $N$ description--word pairs were sampled to form the demonstrations. The remaining concepts were used as the test set. For the other inference tasks, we sampled 24 demonstrations at random from each dataset and evaluated model performance on the remaining examples. For each experimental condition, we repeated evaluation over five independent runs, each with a distinct random sample of $N$ demonstrations.

\subsection{Generalized Canonical Correlation Analysis (GCCA)}
\label{si:methods-gcca}

\paragraph{GCCA Eigenvalue Formulation.}

We employed GCCA to identify a low-dimensional subspace that is shared across model layers. Let $\mathbf{X}_{\ell} \in \mathbb{R}^{n \times d}$ denote the matrix of hidden states at layer $\ell$, where each row $\mathbf{X}_{\ell}[i,:] = \mathbf{h}_{\ell,i}$ corresponds to the representation of the $i^{\textrm{th}}$ query concept, $n$ is the number of query concepts, and $d$ is the hidden dimensionality of the model. Following standard formulations of GCCA~\citep{horst_generalized_1961, kettenring_canonical_1971, benton_deep_2019}, the optimization objective in Equation~\ref{eq:gcca} is equivalent to computing the leading eigenvectors of the aggregate projection operator 
\begin{equation}
\mathbf{S} = \sum_{\ell} \mathbf{P}_{\ell},
\end{equation}
where $\mathbf{P}_{\ell} = \mathbf{X}_{\ell} \left(\mathbf{X}_{\ell}^\top \mathbf{X}_{\ell} + \lambda \mathbf{I}\right)^{-1} \mathbf{X}_{\ell}^\top$ denotes the regularized projection matrix associated with layer $\ell$. Unless otherwise stated, we set the ridge regularization parameter to
$\lambda = 0.01$ for numerical stability. The eigenvectors of $\mathbf{S}$ corresponding to its largest eigenvalues define directions that exhibit strong alignment across layers, and thus constitute the
shared subspace.

\paragraph{Rank Selection via Permutation Testing.}

To determine the dimensionality $\hat{r}$ of the conceptual subspace shared across layers, we employed a non-parametric permutation test. The null hypothesis is that any apparent alignment across layers arises from chance correlations rather than shared structure. Specifically, for each layer $\ell$, we independently applied a random permutation $\pi_{\ell}$ to the rows of $\mathbf{X}_{\ell}$, yielding a permuted matrix $\mathbf{X}_{\ell}^{\pi}$. This procedure destroys the correspondence between representations of the same query concept across layers, while preserving the marginal covariance structure within each layer. We repeated this permutation procedure $M = 500$ times. For each permutation, we computed the eigenvalue spectrum $\{\lambda_i^{\pi}\}$ of the corresponding GCCA operator $\mathbf{S}^{\pi}$. For each component index $i$, we defined a significance threshold $q_i$ as the $(1-\alpha)$ quantile of the null distribution of $\lambda_i^{\pi}$, with $\alpha = 0.05$. The effective rank $\hat{r}$ of the shared subspace was then determined as the number of eigenvalues from the original (unpermuted) data whose magnitude exceeded the corresponding null threshold:
\begin{equation}
    \hat{r} = \sum_{i=1}^{r_{\max}} \mathbb{I}\!\left( \lambda_i > q_i \right).
\end{equation}
To verify that this permutation-based choice of dimensionality was reasonable, we additionally evaluated a range of alternative subspace dimensions, $r \in \left\{64, 128, 256, 512, 1024, 2048\right\}$; the corresponding results are reported in Appendix~\ref{si:results-subspace-dim}.

\subsection{Structural Analysis of Conceptual Subspaces}
\label{si:methods-structure}

\paragraph{GCCA Alignment.}

Given hidden-state matrices $\{\mathbf{X}_{\ell}\}_{\ell \in \mathcal{L}}$ from a set of layers $\mathcal{L}$, GCCA yields a shared latent representation $\mathbf{G} \in \mathbb{R}^{n \times r}$ and layer-specific projection matrices $\mathbf{W}_{\ell} \in \mathbb{R}^{d \times r}$ such that $\mathbf{X}_{\ell}\mathbf{W}_{\ell} \approx \mathbf{G}$. We denote the layer-wise projected representations as $\mathbf{Y}_{\ell} = \mathbf{X}_{\ell}\mathbf{W}_{\ell}$. To quantify the consistency with which different layers express the shared subspace, we computed the correlation between corresponding dimensions of $\mathbf{Y}_{\ell}$ across layers. For a pair of layers $(\ell, \ell')$, GCCA alignment is calculated as
\begin{equation}
    \mathrm{Align}(\ell,\ell') = \frac{1}{r} \sum_{j=1}^{r}
    \mathrm{corr}\!\left(
    \mathbf{Y}_{\ell,:,j}, \mathbf{Y}_{\ell',:,j}
    \right),
\end{equation}
where $\mathbf{Y}_{\ell,:,j}$ denotes the $j$-th column of $\mathbf{Y}_{\ell}$ and $\mathrm{corr}(\cdot,\cdot)$ denotes Pearson correlation. 

\paragraph{Subspace Overlap via Principal Angles.}

To compare subspaces derived under different contexts, we assessed the overlap between the column spaces of the corresponding projection matrices $\mathbf{W}_{\ell}^{c}$, where $c$ indexes the set of demonstrations provided in context. For each $\mathbf{W}_{\ell}^{c}$, we obtained an orthonormal basis $\mathbf{U}_{\ell}^{c}$ via QR decomposition. Given two sets of demonstrations $c$ and $c^{\prime}$, we computed the principal angles $\{\theta_i\}_{i=1}^{r}$ between the subspaces spanned by $\mathbf{U}_{\ell}^{c}$ and $\mathbf{U}_{\ell}^{c^{\prime}}$. Subspace overlap is quantified as the mean squared cosine of these angles: $\frac{1}{r} \sum_{i=1}^{r} \cos^2 \theta_i$. This metric measures whether different contexts induce similar linear subspaces within the model's hidden-state space.

\paragraph{Representational Similarity Analysis (RSA).}

We employed representational similarity analysis (RSA) to compare the relational structure of concepts encoded within different conceptual subspaces. RSA is nonparametric and has been widely adopted to measure alignment between representation spaces based on their similarity structure. For each context (prompt) $s$ and layer $\ell$, we computed a representational dissimilarity matrix (RDM) $\mathbf{D}_{\ell}^{c} \in \mathbb{R}^{n \times n}$ from the projected representations $\mathbf{Y}_{\ell}^{c} = \mathbf{X}_{\ell}^{c} \mathbf{W}_{\ell}^{c}$, where each entry $\mathbf{D}_{\ell,ij}^{c} = 1 - \cos \!\left(\mathbf{Y}_{\ell,i,:}^{c}, \mathbf{Y}_{\ell,j,:}^{c} \right)$
measures the dissimilarity between concepts $i$ and $j$. RSA alignment between representations derived under contexts $s$ and $s^{\prime}$ is then measured by the Spearman's rank correlation between the upper-triangular entries of $\mathbf{D}_{\ell}^{c}$ and $\mathbf{D}_{\ell}^{c^{\prime}}$. 

\subsection{Behavioral Evaluation}
\label{si:methods-behavior}

We evaluated model performance in inference by truncating each generation at the first newline character (``\textbackslash n'') and checking whether the resulting output exactly matched the target word or any of its listed synonyms. We used greedy decoding for all models to ensure a consistent and controlled comparison across model sizes and families.

\subsection{Causal Mediation Analysis}
\label{si:methods-cma}

\paragraph{Causal Indirect Effect and Causal Mediation Score.}

We employed log probabilities to measure the causal effect of interventions on the model's predictions. Specifically, the causal indirect effect (CIE) quantifies the change in the model's output after patching a specific component $\mathbf{h}_{\ell}^{\textrm{patch}}$ at layer $\ell$. We calculated this as the difference in log probabilities for the target (correct) token $y_{q}$ under the intervention, relative to the original (corrupted) prompt $\tilde{s}$:
\begin{equation}
    \mathrm{CIE} = \log f_{\mathcal{M}}\!\left(\tilde{s} \mid \mathbf{h}_{\ell}^{\textrm{patch}} \right)\!\left[y_{q}\right] - \log f_{\mathcal{M}}\!\left(\tilde{s} \right)\!\left[y_{q}\right],
\end{equation}
where $f_{\mathcal{M}}\!\left(\cdot\right)\!\left[y_{q}\right]$ denotes the output probability for token $y_{q}$ from the model $\mathcal{M}$.

For cross-context transfer from a source context $c_{i}$ to a target context $c_{j}$, where $y_{q_b}$ is the expected concept for the query description $s_{q_{b}}$ in the target context, we measured causal mediation (CMA) score. This score represents the change in the log probability difference between the source-related token $y_{q_a}$ and the target-correct token $y_{q_b}$ after the intervention: 
\begin{equation}
\begin{aligned}
    \mathrm{CMA} &= \left(\log f_{\mathcal{M}}\!\left(c_{j}, s_{q_{b}} \mid \mathbf{h}_{\ell}^{\textrm{patch}} \right)\left[y_{q_{a}}\right] - \log f_{\mathcal{M}}\!\left(c_{j}, s_{q_{b}} \mid \mathbf{h}_{\ell}^{\textrm{patch}} \right)\left[y_{q_{b}}\right]\right) \\ & \qquad \quad - \left(\log f_{\mathcal{M}}\! \left(c_{j},s_{q_{b}} \right)\left[y_{q_{a}}\right] - \log f_{\mathcal{M}}\! \left(c_{j},s_{q_{b}} \right)\left[y_{q_{b}}\right] \right).
\end{aligned}
\end{equation}

We preferred log probabilities over raw logits because logits are unconstrained outputs that can vary significantly in scale across different conditions. Log probabilities are normalized via the softmax function, ensuring that the metrics reflect the model's relative confidence in its predictions. This makes them a more stable and interpretable objective for quantifying the functional impact of interventions across varying contexts.

\paragraph{Identification of Significant Attention Heads via Permutation Testing.}

To identify the attention heads that are functionally relevant for inference in LLMs, we performed statistical tests to determine which heads exhibit statistically significant causal effects. We employed a non-parametric permutation test to control the family-wise error rate (FWER) at $\alpha = 0.05$, with the null hypothesis assuming the absence of any real causal effects. For each permutation, we disrupted the relationship between specific heads and the model's output by applying random sign-flipping to the observed causal indirect effects (CIEs). Specifically, each CIE was multiplied by a value sampled uniformly from $\{-1, 1\}$, and this procedure was repeated 5,000 times. To account for multiple comparisons, we constructed a null distribution based on the maximum CIE observed for each permutation. The significance threshold was then defined as the $(1-\alpha)$-quantile of this empirical null distribution. Attention heads were classified as significant if their observed CIE exceeded this threshold. We restricted our analysis to these significant heads to focus on those contributing constructively to the in-context concept inference task.

\begin{table}[tb]
\centering
\caption{Overview of the large language models (LLMs) used in our experiments. ``\#Layers'' denotes the total number of layers in each model, including the embedding layer (indexed as layer 0). ``Sel. Layers (0-indexed)'' lists the layers at which GCCA was applied to derive conceptual subspaces, as determined by a layer-wise SVD analysis. ``Dim.'' denotes the dimensionality of the model's hidden states, and ``Subspace Dim.'' reports the dimensionality of the GCCA-derived conceptual subspaces, selected via permutation testing and summarized as median [range] across five runs. All models are publicly available via Hugging Face~\citep{wolf2019huggingface} (\url{https://huggingface.co}).} 
\label{tab:llms-details}
\begin{tabular*}{\textwidth}{@{\extracolsep\fill}llrrrr}
\toprule
\textbf{Series} & \textbf{Model} & \textbf{\#Layers} & \textbf{Sel. Layers (0-indexed)} & \textbf{Dim.} & \textbf{Subspace Dim.}\\
\midrule
Llama~3.1 & \makecell[l]{\href{https://huggingface.co/meta-llama/Llama-3.1-70B}{\texttt{meta-llama/Llama-3.1-70B}} \\ \href{https://huggingface.co/meta-llama/Llama-3.1-8B}{\texttt{meta-llama/Llama-3.1-8B}}} & \makecell[r]{81 \\ 33} & \makecell[r]{38--79 \\ 16--31} & \makecell[r]{ 8192 \\ 4096 } & \makecell[r]{1180 [1176--1181] \\ 1250 [1242--1254]} \\
\midrule
Llama~3 & \href{https://huggingface.co/meta-llama/Meta-Llama-3-8B}{\texttt{meta-llama/Meta-Llama-3-8B}} & 33 & 16--31 & 4096 & 1245 [1244--1249] \\
\midrule
Qwen2.5 & \makecell[l]{\href{https://huggingface.co/Qwen/Qwen2.5-32B}{\texttt{Qwen/Qwen2.5-32B}} \\ \href{https://huggingface.co/Qwen/Qwen2.5-7B}{\texttt{Qwen/Qwen2.5-7B}} \\ \href{https://huggingface.co/Qwen/Qwen2.5-3B}{\texttt{Qwen/Qwen2.5-3B}}} & \makecell[r]{65 \\ 29 \\ 37} & \makecell[r]{50--63 \\ 22--28 \\ 32--36} & \makecell[r]{5120 \\ 3584 \\ 2048} & \makecell[r]{1221 [1218--1224] \\ 1292 [1290--1299] \\ 1537 [1531--1540]} \\
\noalign{\hrule height 1.2pt}
\end{tabular*}
\end{table}

\begin{table}[tb]
\centering
\caption{Overview of the inference tasks used in our experiments. Data for concept inference were drawn from the THINGS database~\citep{hebart_things_2019} (\url{https://osf.io/jum2f/}); data for the remaining four tasks were taken from \citet{todd_function_2024} (\url{https://github.com/ericwtodd/function_vectors}).} 
\label{tab:task-details}
\begin{tabular*}{\textwidth}{@{\extracolsep\fill}lll}
\toprule
\textbf{Task} & \textbf{Examples} & \textbf{Source} \\
\midrule
Concept Inference & \makecell[l]{a small very thin pancake $\Rightarrow$ crepe \\ a small guitar having four strings $\Rightarrow$ ukulele \\ dried grape $\Rightarrow$ raisin} & \citet{hebart_things_2019} \\
\midrule
Antonym & \makecell[l]{true $\Rightarrow$ false \\ difficult $\Rightarrow$ easy \\ proceed $\Rightarrow$ halt} & \citet{nguyen_distinguishing_2017} \\
\midrule
Country--Capital & \makecell[l]{Austria $\Rightarrow$ Vienna \\ China $\Rightarrow$ Beijing \\ Germany $\Rightarrow$ Berlin} & \citet{todd_function_2024} \\
\midrule
Landmark--Country & \makecell[l]{Achelous River $\Rightarrow$ Greece \\ Atwell Peak $\Rightarrow$ Canada \\ Beauvais $\Rightarrow$ France} & \citet{hernandez2024linearity} \\
\midrule
National Parks & \makecell[l]{Prince William Forest Park $\Rightarrow$ Virginia \\ Hot Springs National Park $\Rightarrow$ Arkansas \\ Kings Canyon National Park $\Rightarrow$ California} & \citet{todd_function_2024} \\
\noalign{\hrule height 1.2pt} 
\end{tabular*}
\end{table}

\begin{figure*}[t]
    \centering
    
    \textbf{a. Description corruption}\par
    \includegraphics[width=0.75\textwidth]{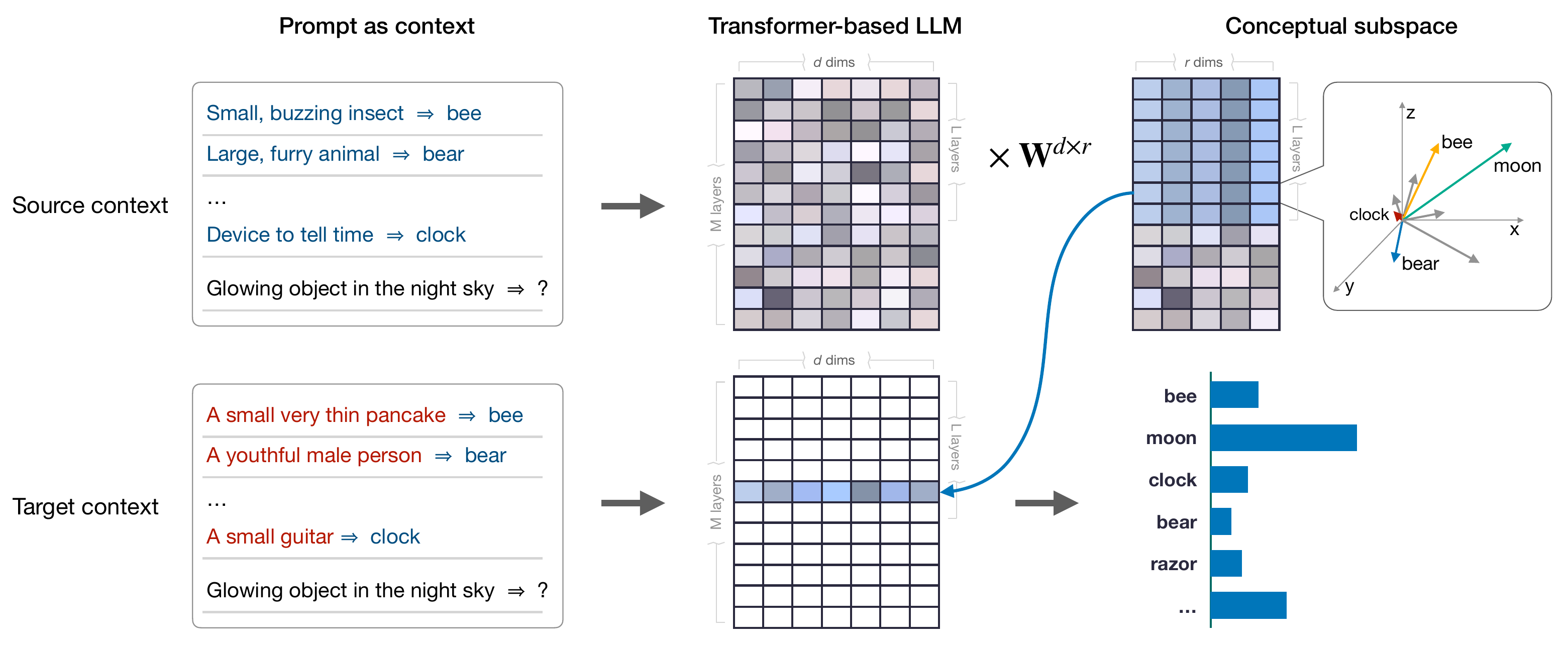}
    
    \vspace{0.8em}
    
    \textbf{b. Label corruption}\par
    \includegraphics[width=0.75\textwidth]{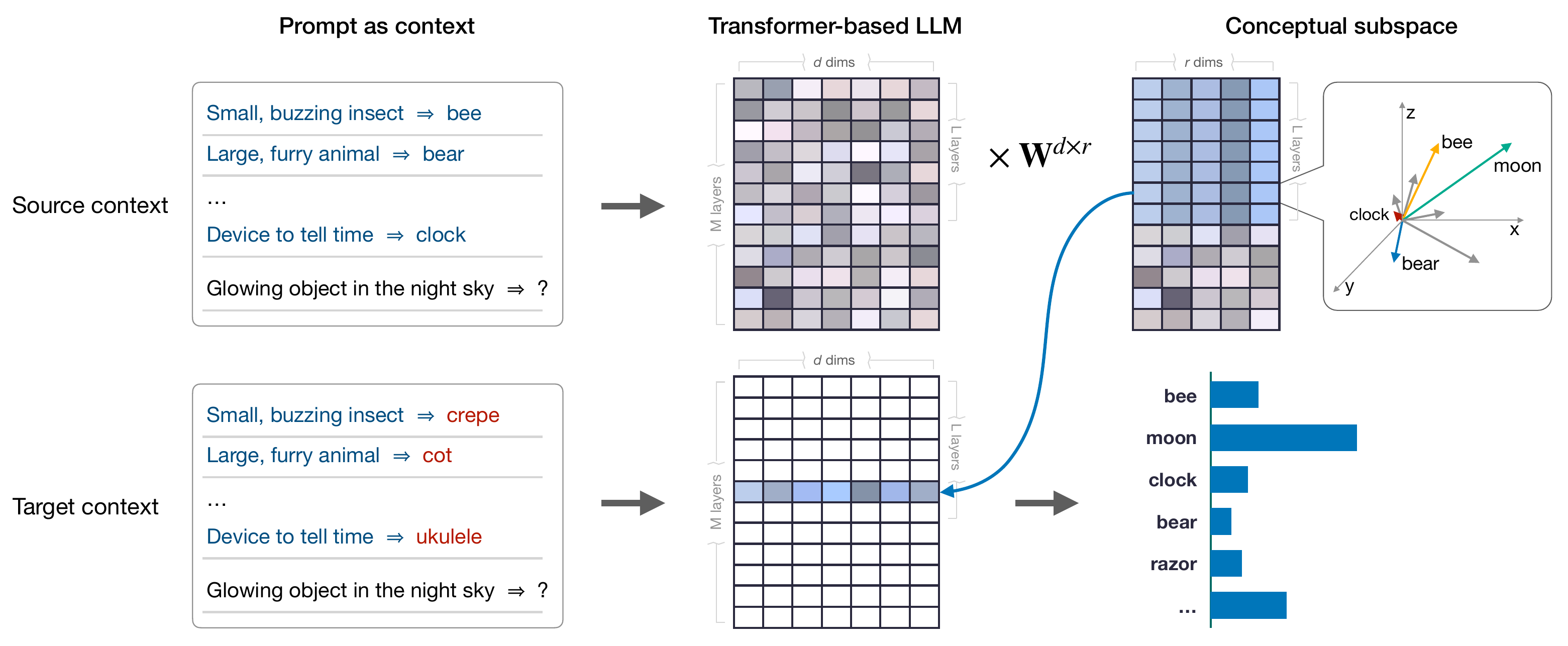}
    
    \vspace{0.8em}
    
    \textbf{c. Query corruption}\par
    \includegraphics[width=0.75\textwidth]{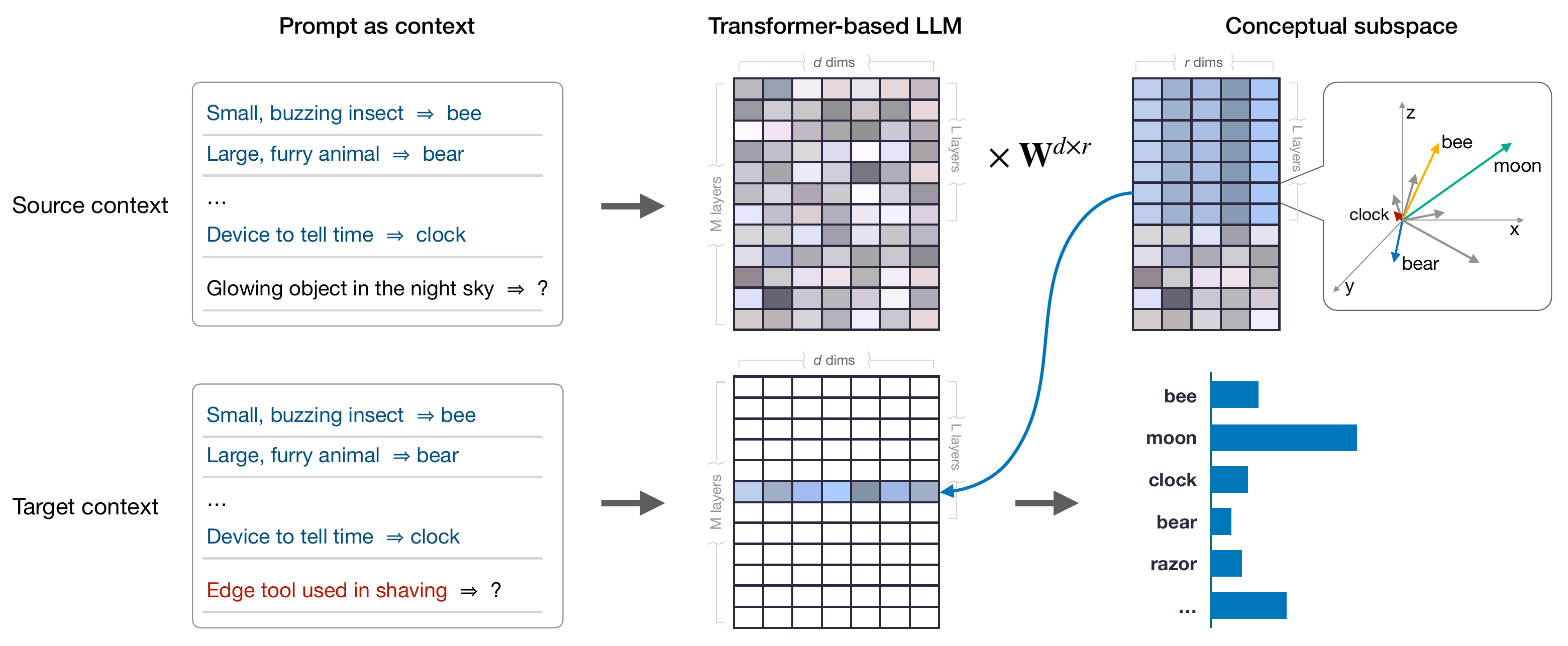}
    
    \caption{Illustration of causal mediation analysis under three corruption conditions: description (\textbf{a}), label (\textbf{b}), and query (\textbf{c}) corruption. In the source context, the prompt contains correct description--word pairs as demonstrations, together with a query description. In the target context, the corrupted field is replaced with mismatched content of the same token length. We then patch the conceptual subspace from the source context into the target context to test whether the model's prediction shifts systematically toward the expected source-context term. The figure illustrates this procedure using the concept inference task as an example.}
    \label{fig:si-corruption}
\end{figure*}

\clearpage

\section{Supplementary Results}

\subsection{Evaluating Subspace Dimensionality}
\label{si:results-subspace-dim}

\begin{figure}[t]
\begin{center}
\centerline{\includegraphics[width=\linewidth]{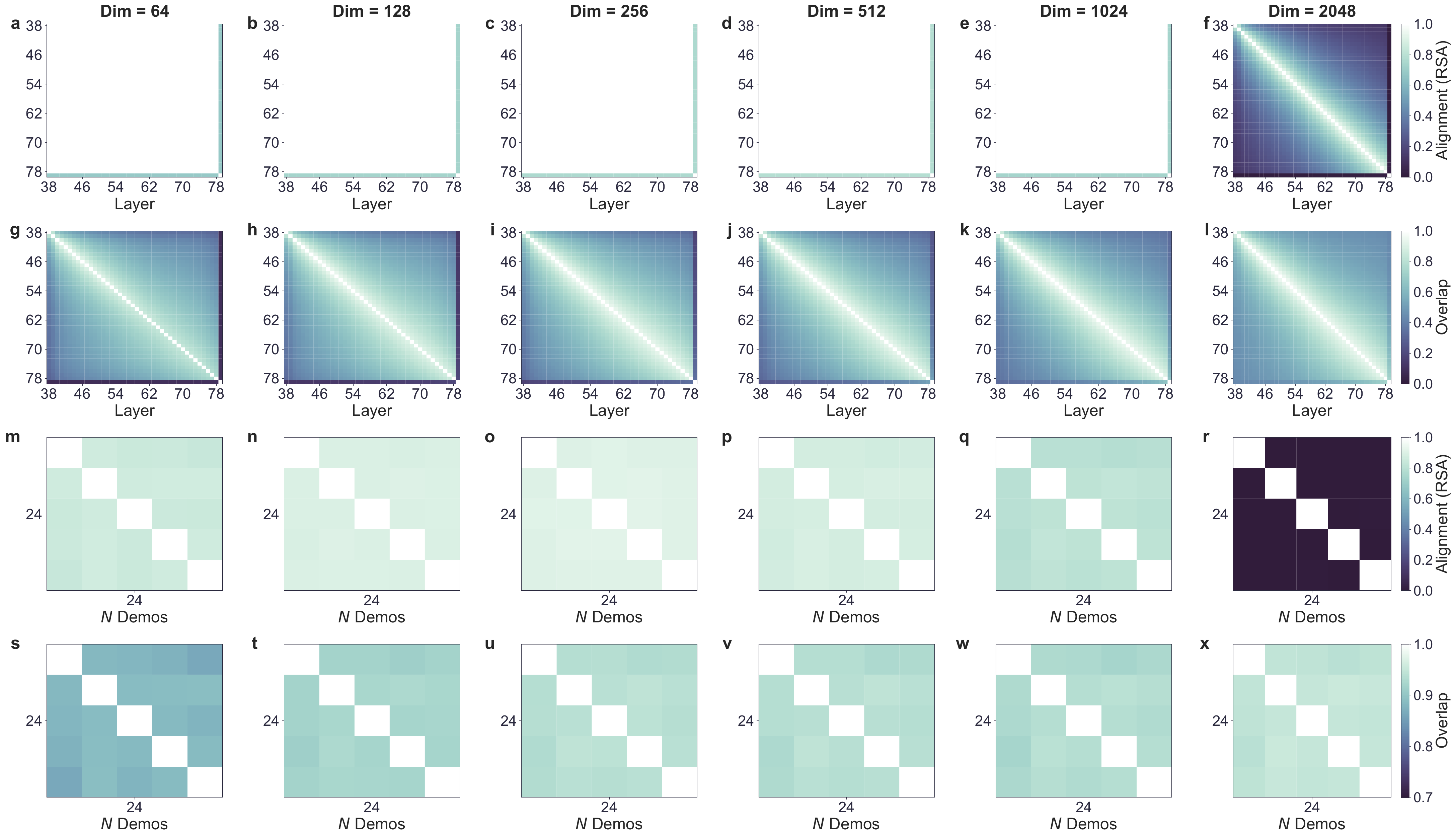}}
\caption{Identifying shared subspaces across layers in Llama-3.1~70B at varying dimensionalities using GCCA. Titles indicate the retained subspace dimensionality. All results are computed with 24 in-context demonstrations and averaged over five runs. \textbf{a}--\textbf{f}, Alignment of representational geometry between GCCA-derived subspace across selected layers, measured by RSA, with both axes indexing layers. \textbf{g}--\textbf{l}, Overlap between GCCA-derived projection matrices across selected layers, with both axes indexing layers. \textbf{m}--\textbf{r}, Cross-context alignment of representational geometry within the GCCA-derived subspace across different demonstration sets, measured by RSA. \textbf{s}--\textbf{x}, Overlap between GCCA-derived projection matrices across demonstration sets. For \textbf{m}--\textbf{x}, the axes denote the number of demonstrations, and each matrix entry corresponds to a single run.}\label{fig:si-subspace_dim}
\end{center}
\end{figure}

To evaluate whether the permutation-based choice of dimensionality ($\hat{r} = 1180$ for Llama-3.1~70B) was well chosen, we additionally tested alternative subspace dimensions, $r \in \left\{64, 128, 256, 512, 1024, 2048\right\}$. For dimensions smaller than $\hat{r}$, the representational results were qualitatively similar: cross-layer alignment remained close to unity (Figure~\ref{fig:si-subspace_dim}a--e), and the GCCA-derived projection matrices varied smoothly across layers (Figure~\ref{fig:si-subspace_dim}g--k). With 24 demonstrations, these lower-dimensional subspaces also exhibited relatively strong alignment in their internal structure (Figure~\ref{fig:si-subspace_dim}m--q) and substantial overlap (Figure~\ref{fig:si-subspace_dim}s--w). In contrast, increasing the dimensionality to 2048 led to markedly weaker cross-layer alignment (Figure~\ref{fig:si-subspace_dim}f) and substantially reduced cross-context alignment (Figure~\ref{fig:si-subspace_dim}r), suggesting that the larger subspace captures additional directions beyond the core shared structure, including information that is less consistently shared and more context-dependent. 

\begin{figure}[t]
\begin{center}
\centerline{\includegraphics[width=\linewidth]{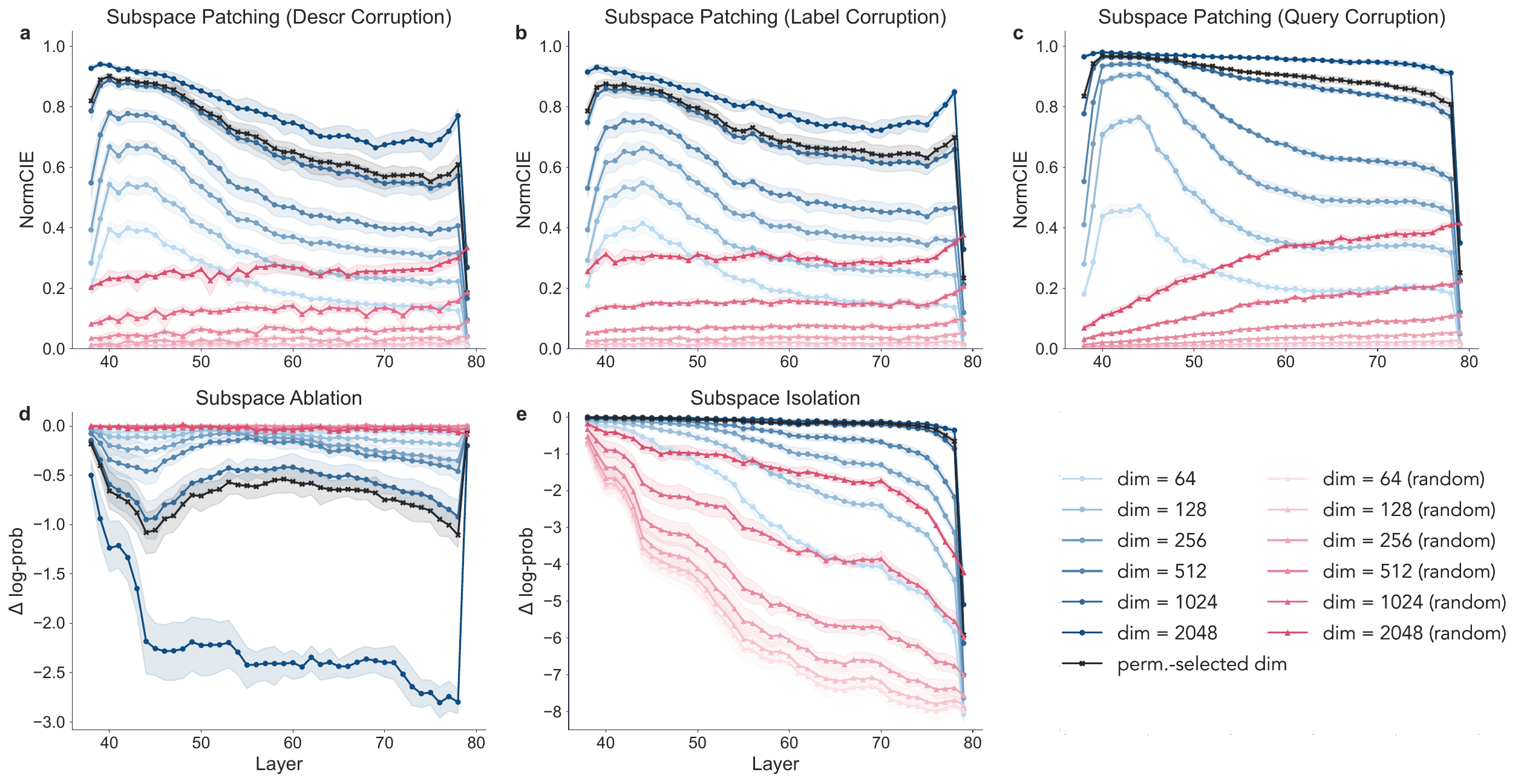}}
\caption{Causal intervention results for Llama-3.1~70B with GCCA-derived subspaces of varying dimensionality. All results are obtained with 24 in-context demonstrations. Results for the GCCA-derived subspace (blue) are compared with dimension-matched random-subspace baselines (red). Black lines denote the permutation-selected dimensionality used in the main text. \textbf{a}--\textbf{c}, Activation patching under three corruption conditions: description (\textbf{a}), label (\textbf{b}) and query (\textbf{c}). The x-axis indexes layers, and the y-axis shows the normalized causal indirect effect (CIE). \textbf{d}--\textbf{e}, Subspace necessity and sufficiency, assessed by ablating the subspace (\textbf{d}) or isolating it (\textbf{e}). The y-axis denotes the change in log-probability of the correct token. Shaded regions indicate 95\% CIs computed from 10,000 bootstrap resamples across five runs.}\label{fig:si-subspace_dim-causality}
\end{center}
\end{figure}

Causal intervention analyses showed progressively stronger effects as subspace dimensionality increased. Under corrupted context, patching the GCCA-derived subspace restored a larger portion of performance as more dimensions were retained (Figure~\ref{fig:si-subspace_dim-causality}a--c), while consistently outperforming random-subspace baselines of matched dimensionality. Although $r=2048$ restored more performance than the permutation-selected $\hat{r}$, the representational analyses suggest that subspaces of this size include additional weaker and more context-dependent directions that may nevertheless remain task-relevant. Ablating the GCCA-derived subspaces (Figure~\ref{fig:si-subspace_dim-causality}d) produced significantly larger performance drops than ablating matched random subspaces, whose effects remained small even at $r=2048$. These ablation effects also became more pronounced with increasing dimensionality. Likewise, isolating the GCCA-derived subspaces preserved substantially more performance than matched random-subspace baselines (Figure~\ref{fig:si-subspace_dim-causality}e). Once the dimensionality approached the permutation-selected $\hat{r}$, isolation preserved near-original performance across most layers, and the effect was close to that obtained with $r=2048$. Importantly, across all tested dimensions and intervention settings, the conceptual subspace at the permutation-selected dimensionality consistently outperformed random baselines by a wide margin, supporting the validity of the selected rank.

Together, these results indicate that the permutation-selected dimensionality is a reasonable operating point: it captures the relational structure shared across layers and contexts while retaining most of the task-relevant information functionally used by the model. Increasing the dimensionality yields stronger patching and ablation effects, but at the cost of incorporating information that is less consistently shared and more context-dependent. This pattern suggests that dimensions beyond the permutation-selected rank may contain complementary task-relevant information used by LLMs, an interesting direction for future investigation.

\subsection{Assessing the Generality of Conceptual Subspaces Across Inference Tasks}
\label{si:results-task-generality}

\begin{figure*}[t]
    \begin{center}
    \centerline{\includegraphics[width=\textwidth]{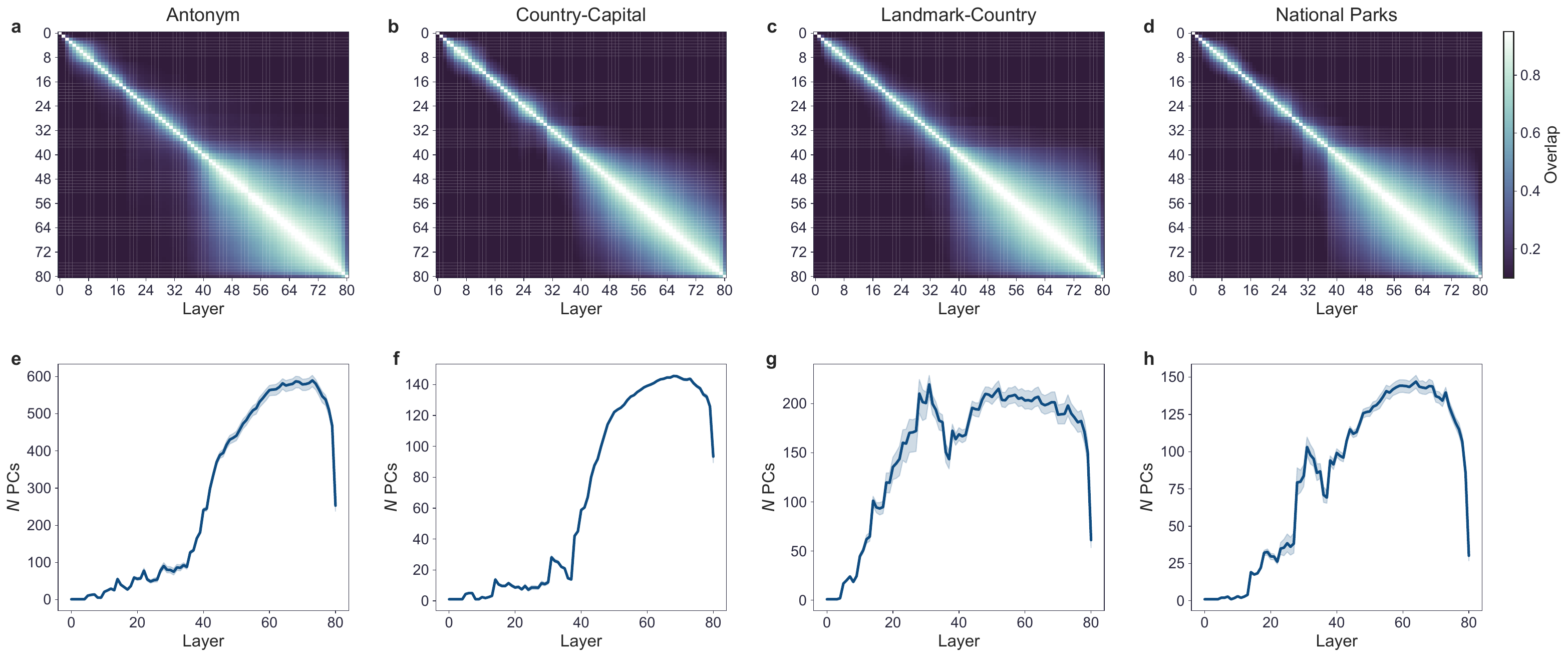}}
    \caption{Emergence of conceptual subspaces in Llama-3.1~70B across four in-context inference tasks. \textbf{a}--\textbf{d}, Layer-wise similarity of hidden states, measured as subspace overlap (mean squared cosine of principal angles) between SVD subspaces explaining 95\% variance. Results are averaged over five runs, each with 24 demonstrations. Axes index layers. \textbf{e}--\textbf{h}, Number of principal components (PCs) needed to explain 95\% of the variance at each layer. Shaded areas denote 95\% CIs calculated from 10,000 bootstrap resamples over five runs.}
    \label{fig:si-subspace_emerge-simple_tasks}
    \end{center}
\end{figure*}

To assess whether the identified conceptual subspaces generalize beyond concept inference, we extended our analyses to additional inference tasks from \citet{todd_function_2024}, including Antonym, Country--Capital, Landmark--Country, and National Parks (Table~\ref{tab:task-details}). Across these tasks, layer-wise representational analyses revealed the same qualitative pattern observed in the concept inference setting: shared subspaces emerged in the middle layers (approximately layers 38--40) and persisted into later layers (Figure~\ref{fig:si-subspace_emerge-simple_tasks}a--d). The emergence of these subspaces was likewise accompanied by an increase in the number of principal components (PCs) required to explain 95\% of the variance (Figure~\ref{fig:si-subspace_emerge-simple_tasks}e--h), suggesting that contextual information is integrated into a higher-dimensional representational space for subsequent computation. Together, these findings mirror those observed for concept inference (Section~\ref{sec:emergence-results}), while also indicating task-dependent variation in dimensionality, which may reflect differences in task complexity.

\begin{figure*}[t]
    \begin{center}
    \centerline{\includegraphics[width=\textwidth]{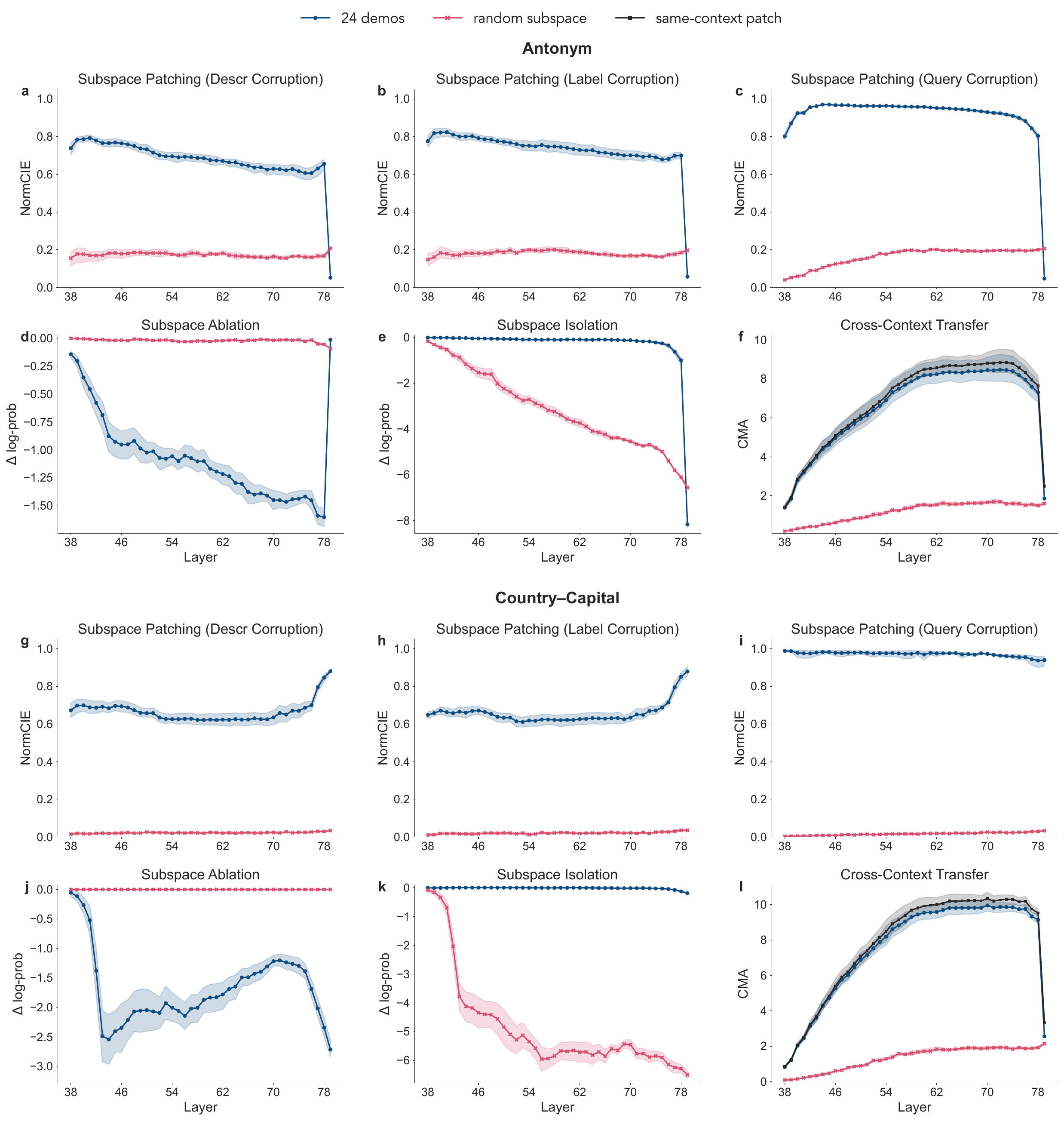}}
    \caption{Causal intervention results for Llama-3.1~70B on two in-context inference tasks (Antonym and Country--Capital). \textbf{a}--\textbf{c}, \textbf{g}--\textbf{i}, Activation patching with 24 demonstrations under three corruption conditions: description (\textbf{a}, \textbf{g}), label (\textbf{b}, \textbf{h}) and query (\textbf{c}, \textbf{i}). The conceptual subspace (blue) is compared against a random-subspace baseline (red). The x-axis indexes layers, and the y-axis shows the normalized causal indirect effect (CIE). \textbf{d}--\textbf{e}, \textbf{j}--\textbf{k}, Subspace necessity and sufficiency tested by ablating the conceptual subspace (\textbf{d}, \textbf{j}) or isolating it (\textbf{e}, \textbf{k}). The y-axis denotes the change in log-probability of the correct token. \textbf{f}, \textbf{l}, Causal effects of cross-context transfer, where the relational structure from a source context is adapted to a target context via an orthogonal transformation. The y-axis reports the causal mediation (CMA) score. Performance is compared against a random-subspace baseline (red). Shaded regions indicate 95\% CIs computed from 10,000 bootstrap resamples across five runs.}
    \label{fig:si-causality-simple_tasks-antonym}
    \end{center}
\end{figure*}

\begin{figure*}[t]
    \begin{center}
    \centerline{\includegraphics[width=\textwidth]{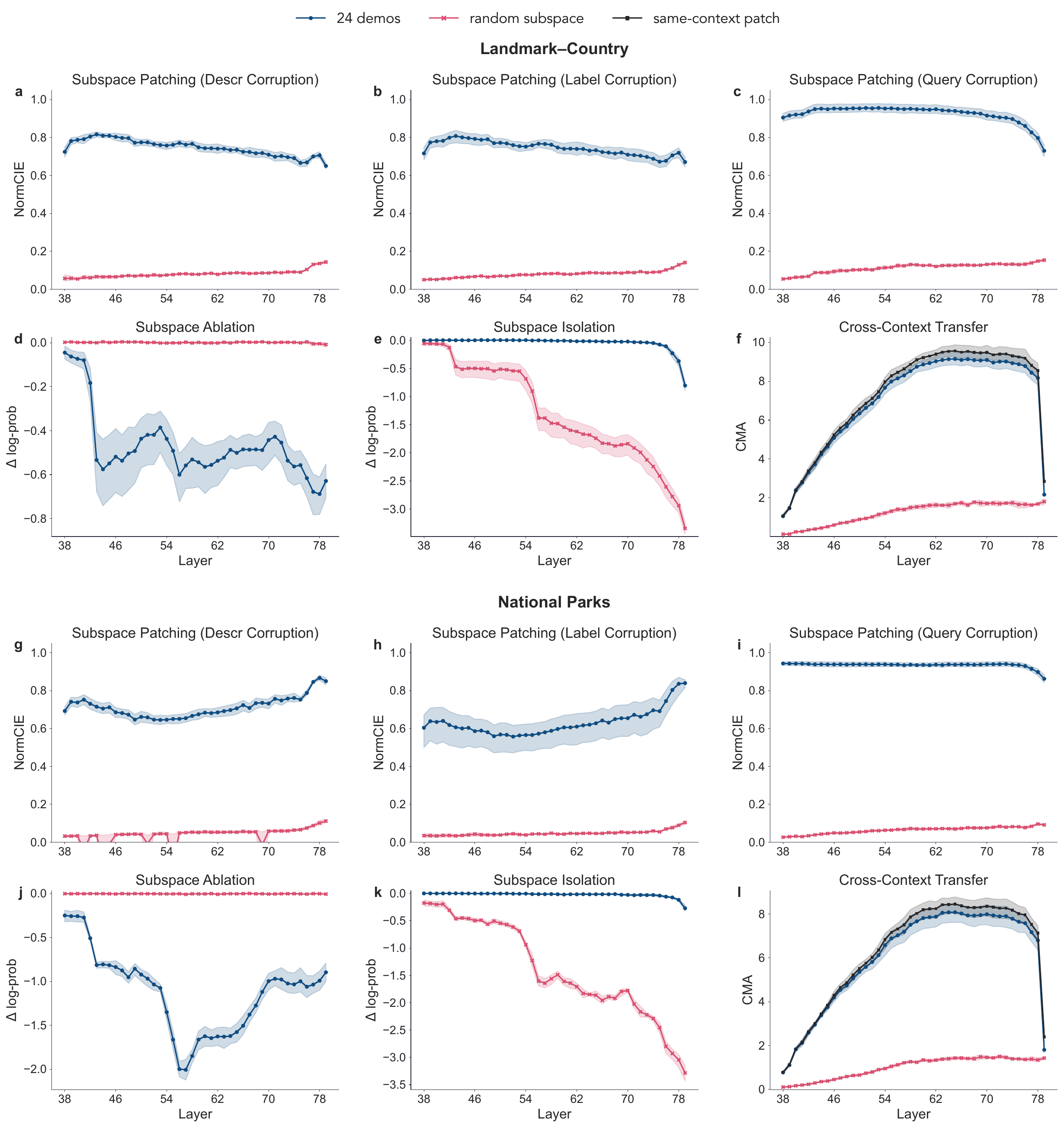}}
    \caption{Causal intervention results for Llama-3.1~70B on two in-context inference tasks (Landmark--Country and National Parks). \textbf{a}--\textbf{c}, \textbf{g}--\textbf{i}, Activation patching with 24 demonstrations under three corruption conditions: description (\textbf{a}, \textbf{g}), label (\textbf{b}, \textbf{h}) and query (\textbf{c}, \textbf{i}). The conceptual subspace (blue) is compared against a random-subspace baseline (red). The x-axis indexes layers, and the y-axis shows the normalized causal indirect effect (CIE). \textbf{d}--\textbf{e}, \textbf{j}--\textbf{k}, Subspace necessity and sufficiency tested by ablating the conceptual subspace (\textbf{d}, \textbf{j}) or isolating it (\textbf{e}, \textbf{k}). The y-axis denotes the change in log-probability of the correct token. \textbf{f}, \textbf{l}, Causal effects of cross-context transfer, where the relational structure from a source context is adapted to a target context via an orthogonal transformation. The y-axis reports the causal mediation (CMA) score. Performance is compared against a random-subspace baseline (red). Shaded regions indicate 95\% CIs computed from 10,000 bootstrap resamples across five runs.}
    \label{fig:si-causality-simple_tasks-landmark_country}
    \end{center}
\end{figure*}

Causal interventions revealed a similar pattern across tasks. Under corrupted context, performance was largely restored by patching the conceptual subspace into middle layers, substantially outperforming the random-subspace baseline (Figures~\ref{fig:si-causality-simple_tasks-antonym}a--c,g--i; \ref{fig:si-causality-simple_tasks-landmark_country}a--c,g--i). Ablation and isolation analyses likewise supported the necessity and sufficiency of the subspace for inference (Figures~\ref{fig:si-causality-simple_tasks-antonym}d--e,j--k; \ref{fig:si-causality-simple_tasks-landmark_country}d--e,j--k): ablating the conceptual subspace caused a marked performance drop relative to the random-subspace baseline, whereas isolating it preserved near-original performance across most layers. In cross-context transfer, patching effects closely approached the same-context reference and remained well above the random-subspace baseline (Figures~\ref{fig:si-causality-simple_tasks-antonym}f,i; \ref{fig:si-causality-simple_tasks-landmark_country}f,i), suggesting that the model constructs and exploits a stable, abstract conceptual subspace whose relational structure is preserved across contexts. Overall, these results indicate that the emergence and functional role of conceptual subspaces are not unique to concept inference, but extend across a broader range of in-context inference tasks, although the precise layers at which the subspace mediates contextual information or supports cross-context abstraction may differ by task.

\subsection{Identifying Conceptual Subspaces across Models}
\label{si:results-emergence}

\begin{figure*}[t]
    \begin{center}
    \centerline{\includegraphics[width=0.9\textwidth]{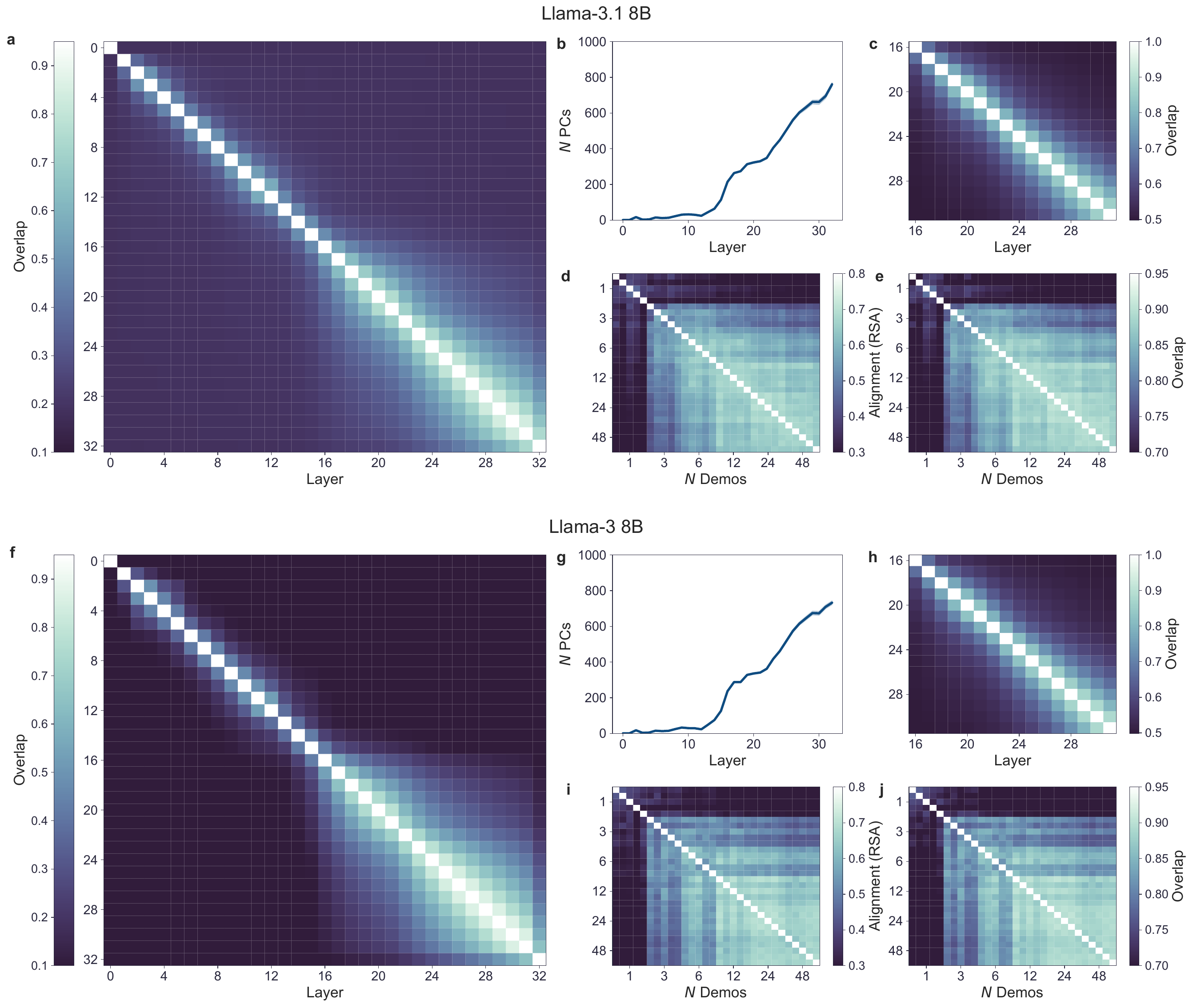}}
    \caption{Emergence of conceptual subspaces in Llama-3.1~8B (\textbf{a}--\textbf{e}) and Llama-3~8B (\textbf{f}--\textbf{j}). \textbf{a}, \textbf{f}, Layer-wise similarity of hidden states, measured as subspace overlap (mean squared cosine of principal angles) between SVD subspaces explaining 95\% variance; results are averaged over five runs with 24 demonstrations. Axes index layers. \textbf{b}, \textbf{g}, Number of principal components (PCs) required to explain 95\% of the variance across layers. Shaded areas denote 95\% CIs calculated from 10,000 bootstrap resamples across five runs. \textbf{c}, \textbf{h}, Overlap between GCCA-derived projection matrices across selected layers. \textbf{d}--\textbf{e}, \textbf{i}--\textbf{j}, Conceptual subspaces becomes increasingly stable as the number of in-context demonstrations grows. \textbf{d}, \textbf{i}, Cross-context alignment of representational geometry within the GCCA subspace across demonstration sets, measured by RSA. \textbf{e}, \textbf{j}, Overlap between GCCA-derived projection matrices across demonstration sets. Axes in \textbf{d}--\textbf{e} and \textbf{i}--\textbf{j} indicate the number of demonstrations, with each cell corresponding to a single run.}
    \label{fig:si-subspace_emerge-llama3}
    \end{center}
\end{figure*}

\begin{figure*}[t]
    \begin{center}
    \centerline{\includegraphics[width=0.89\textwidth]{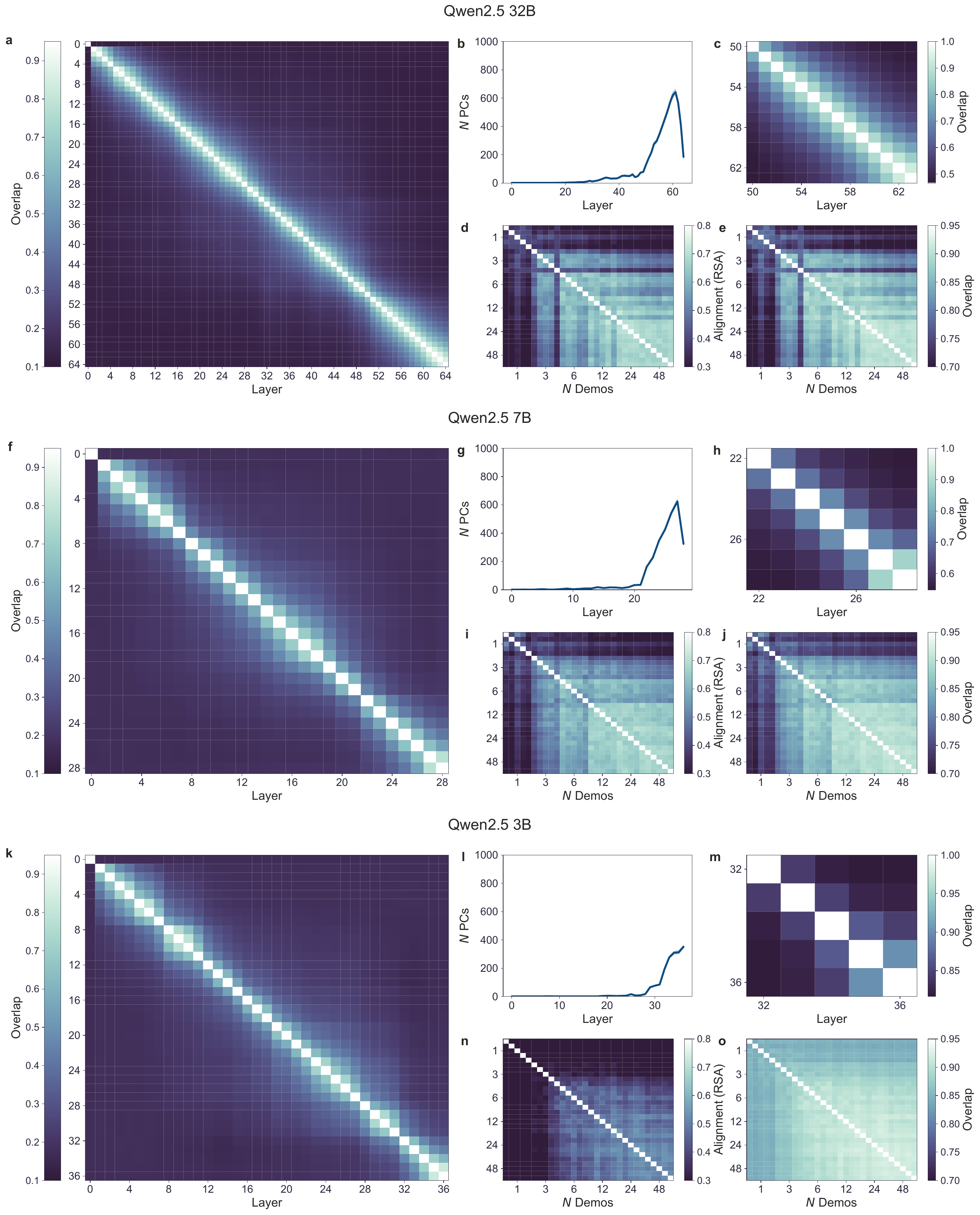}}
    \caption{Emergence of conceptual subspaces in Qwen2.5~32B (\textbf{a}--\textbf{e}), Qwen2.5~7B (\textbf{f}--\textbf{j}), and Qwen2.5~3B (\textbf{k}--\textbf{o}). \textbf{a}, \textbf{f}, \textbf{k}, Layer-wise similarity of hidden states, measured as subspace overlap (mean squared cosine of principal angles) between SVD subspaces explaining 95\% variance; results are averaged over five runs with 24 demonstrations. Axes index layers. \textbf{b}, \textbf{g}, \textbf{l}, Number of principal components (PCs) required to explain 95\% of the variance across layers. Shaded areas denote 95\% CIs calculated from 10,000 bootstrap resamples across five runs. \textbf{c}, \textbf{h}, \textbf{m}, Overlap between GCCA-derived projection matrices across selected layers. \textbf{d}--\textbf{e}, \textbf{i}--\textbf{j}, \textbf{n}--\textbf{o}, Conceptual subspaces becomes increasingly stable as the number of in-context demonstrations grows. \textbf{d}, \textbf{i}, \textbf{n}, Cross-context alignment of representational geometry within the GCCA subspace across demonstration sets, measured by RSA. \textbf{e}, \textbf{j}, \textbf{o}, Overlap between GCCA-derived projection matrices across demonstration sets. Axes in \textbf{d}--\textbf{e}, \textbf{i}--\textbf{j} and \textbf{n}--\textbf{o} indicate the number of demonstrations, with each cell corresponding to a single run.}
    \label{fig:si-subspace_emerge-qwen2v5}
    \end{center}
\end{figure*}

Across all LLMs examined, a shared latent subspace consistently emerged in the middle-to-late layers (Figures~\ref{fig:si-subspace_emerge-llama3}--\ref{fig:si-subspace_emerge-qwen2v5}). Layer-wise SVD showed that early layers exhibited relatively low cross-layer overlap, whereas middle-to-late layers maintained high inter-layer overlap. This transition coincided with a sharp increase in the number of principal components required to explain 95\% of the variance (Figures~\ref{fig:si-subspace_emerge-llama3}a--b,f--g; \ref{fig:si-subspace_emerge-qwen2v5}a--b,f--g,k--l), suggesting the integration of contextual information into a higher-dimensional subspace.

Subspaces identified via GCCA demonstrated near-unity GCCA alignment scores and RSA values across layers. Increasing the number of demonstrations from 1 to 24 led to progressively stronger cross-context alignment of the internal subspace structure, with diminishing gains beyond this threshold (Figures~\ref{fig:si-subspace_emerge-llama3}d--e,i--j; \ref{fig:si-subspace_emerge-qwen2v5}d--e,i--j,n--o). These trends were consistent across model families, though the relative depth at which the shared subspace emerges varied; for example, Qwen2.5 models manifested the shared subspaces at relatively later layers than Llama-3 and Llama-3.1 variants. 

Collectively, these results suggest that LLMs construct a conceptual subspace during inference that persists across middle to late layers. While larger models (e.g., Llama-3.1~70B) required higher-dimensional layer-wise subspaces to capture 95\% of the variance (Figure~\ref{fig:subspace_emerge}b), permutation testing identified fewer significant dimensions for GCCA analyses across layers (Table~\ref{tab:llms-details}). Furthermore, these subspaces exhibited stronger cross-context alignment compared to smaller models. These findings indicate that increased model scale and compute support the formation of richer and more stable relational structures, which are encoded in a more coherent and compressed manner, revealing systematic differences in how LLMs organize internal computations during inference.

\subsection{Analyzing the Causal Role of Conceptual Subspaces in Inference across Models}
\label{si:results-causality}

\begin{figure*}[t]
    \begin{center}
    \centerline{\includegraphics[width=0.9\textwidth]{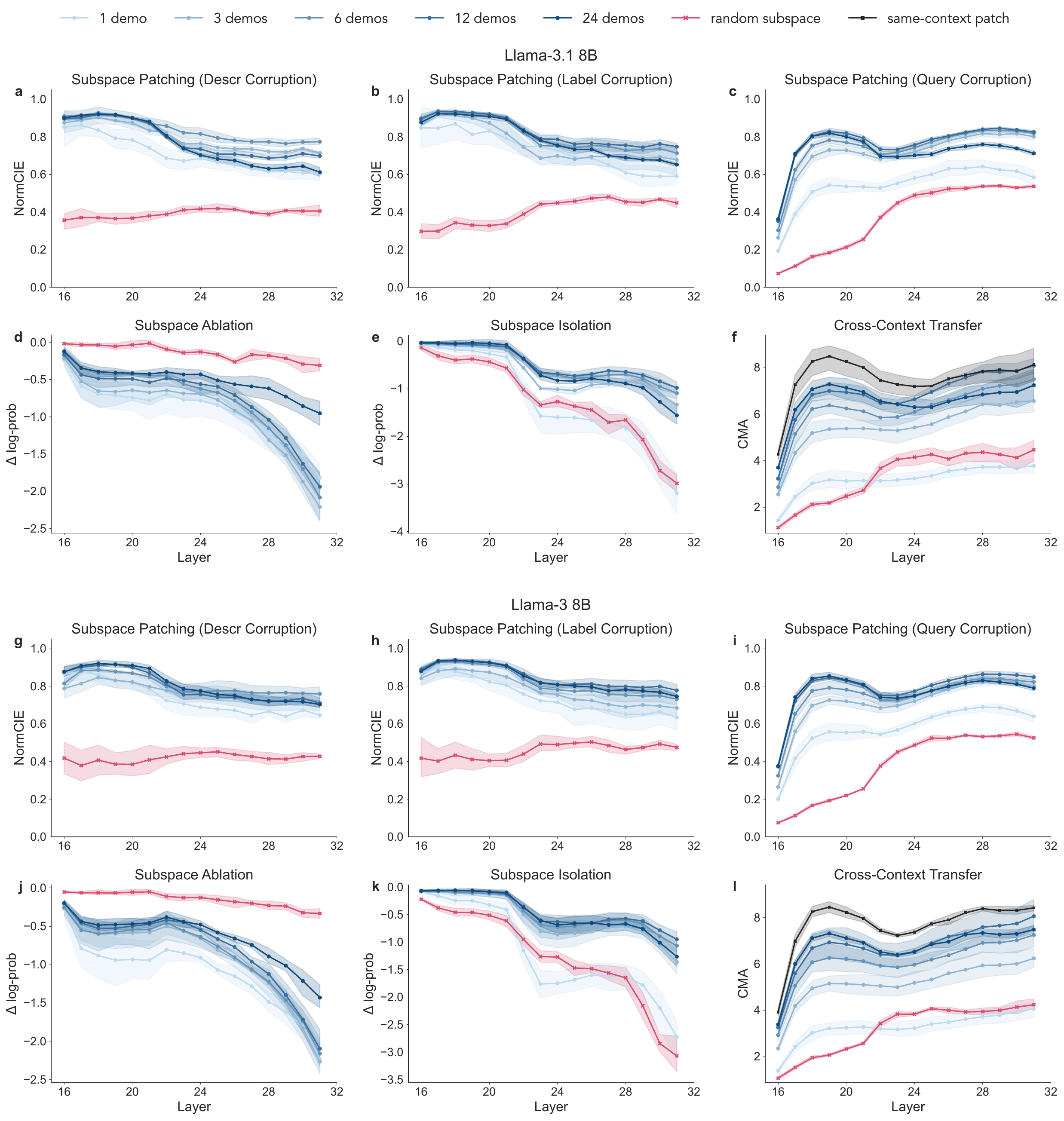}}
    \caption{Causal intervention results for Llama-3.1~8B (\textbf{a}--\textbf{f}) and Llama-3~8B (\textbf{g}--\textbf{l}). \textbf{a}--\textbf{c}, \textbf{g}--\textbf{i}, Activation patching with $N$ demonstrations under three corruption conditions: description (\textbf{a}, \textbf{g}), label (\textbf{b}, \textbf{h}) and query (\textbf{c}, \textbf{i}). Patching the conceptual subspace (blue) is compared against a random-subspace baseline (red). The x-axis indexes layers, and the y-axis shows the normalized causal indirect effect (CIE). \textbf{d}--\textbf{e}, \textbf{j}--\textbf{k}, Subspace necessity and sufficiency tested by ablating the conceptual subspace (\textbf{d}, \textbf{j}) or isolating it (\textbf{e}, \textbf{k}). The y-axis denotes the change in log-probability of the correct token. \textbf{f}, \textbf{l}, Causal effects of cross-context transfer, where the relational structure from a source context is adapted to a target context via an orthogonal transformation. The y-axis reports the causal mediation (CMA) score. Performance is compared against a random-subspace baseline (red) and a same-context patch reference (black) using the conceptual subspace derived from the target context itself. Shaded regions indicate 95\% CIs computed from 10,000 bootstrap resamples across five runs.}
    \label{fig:si-causality-llama3}
    \end{center}
\end{figure*}

\begin{figure*}[t]
    \begin{center}
    \centerline{\includegraphics[width=0.9\textwidth]{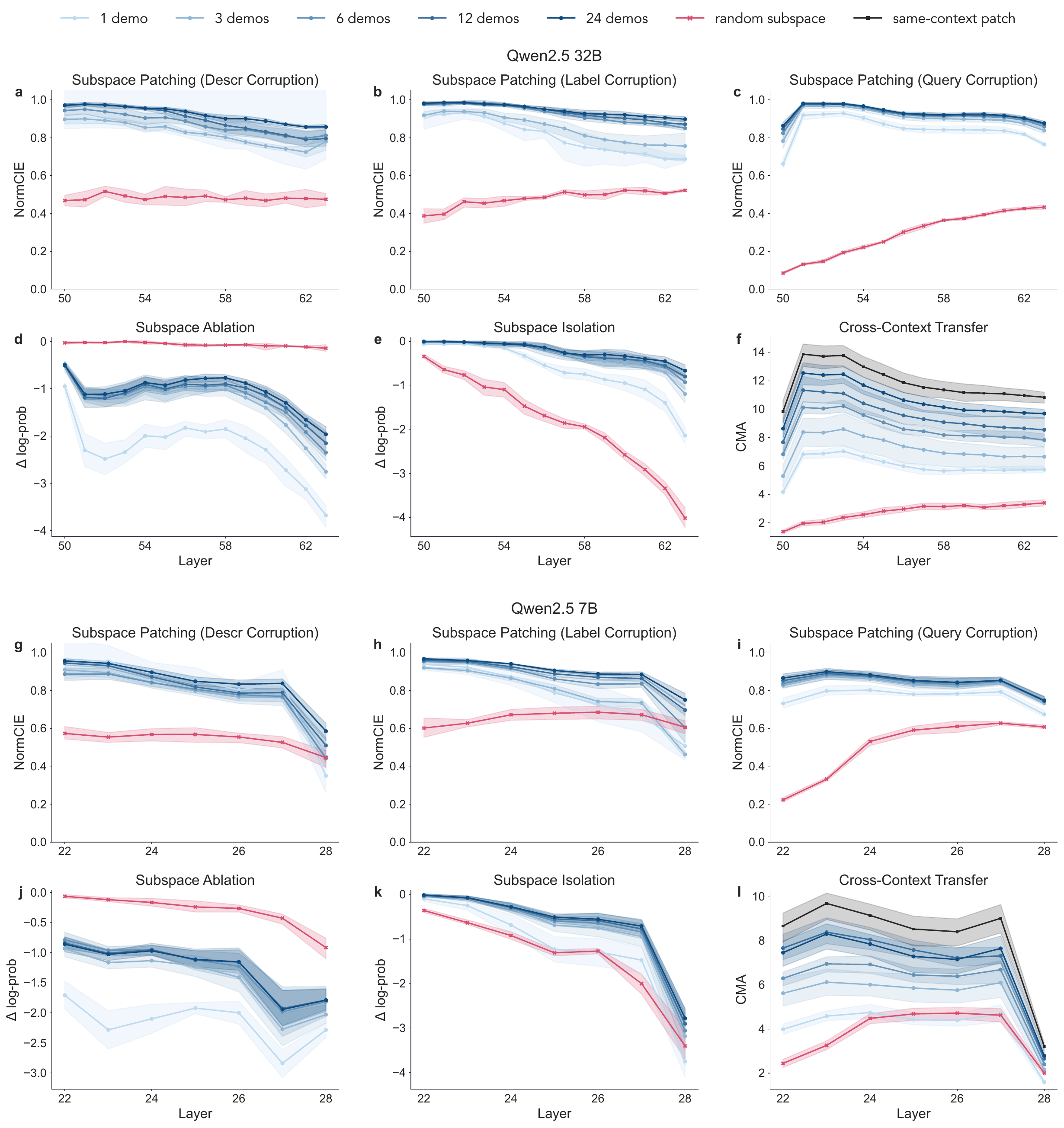}}
    \caption{Causal intervention results for Qwen2.5~32B (\textbf{a}--\textbf{f}) and Qwen2.5~7B (\textbf{g}--\textbf{l}). \textbf{a}--\textbf{c}, \textbf{g}--\textbf{i}, Activation patching with $N$ demonstrations under three corruption conditions: description (\textbf{a}, \textbf{g}), label (\textbf{b}, \textbf{h}) and query (\textbf{c}, \textbf{i}). Patching the conceptual subspace (blue) is compared against a random-subspace baseline (red). The x-axis indexes layers, and the y-axis shows the normalized causal indirect effect (CIE). \textbf{d}--\textbf{e}, \textbf{j}--\textbf{k}, Subspace necessity and sufficiency tested by ablating the conceptual subspace (\textbf{d}, \textbf{j}) or isolating it (\textbf{e}, \textbf{k}). The y-axis denotes the change in log-probability of the correct token. \textbf{f}, \textbf{l}, Causal effects of cross-context transfer, where the relational structure from a source context is adapted to a target context via an orthogonal transformation. The y-axis reports the causal mediation (CMA) score. Performance is compared against a random-subspace baseline (red) and a same-context patch reference (black) using the conceptual subspace derived from the target context itself. Shaded regions indicate 95\% CIs computed from 10,000 bootstrap resamples across five runs.}
    \label{fig:si-causality-qwen2v5}
    \end{center}
\end{figure*}

\begin{figure*}[t]
    \begin{center}
    \centerline{\includegraphics[width=0.9\textwidth]{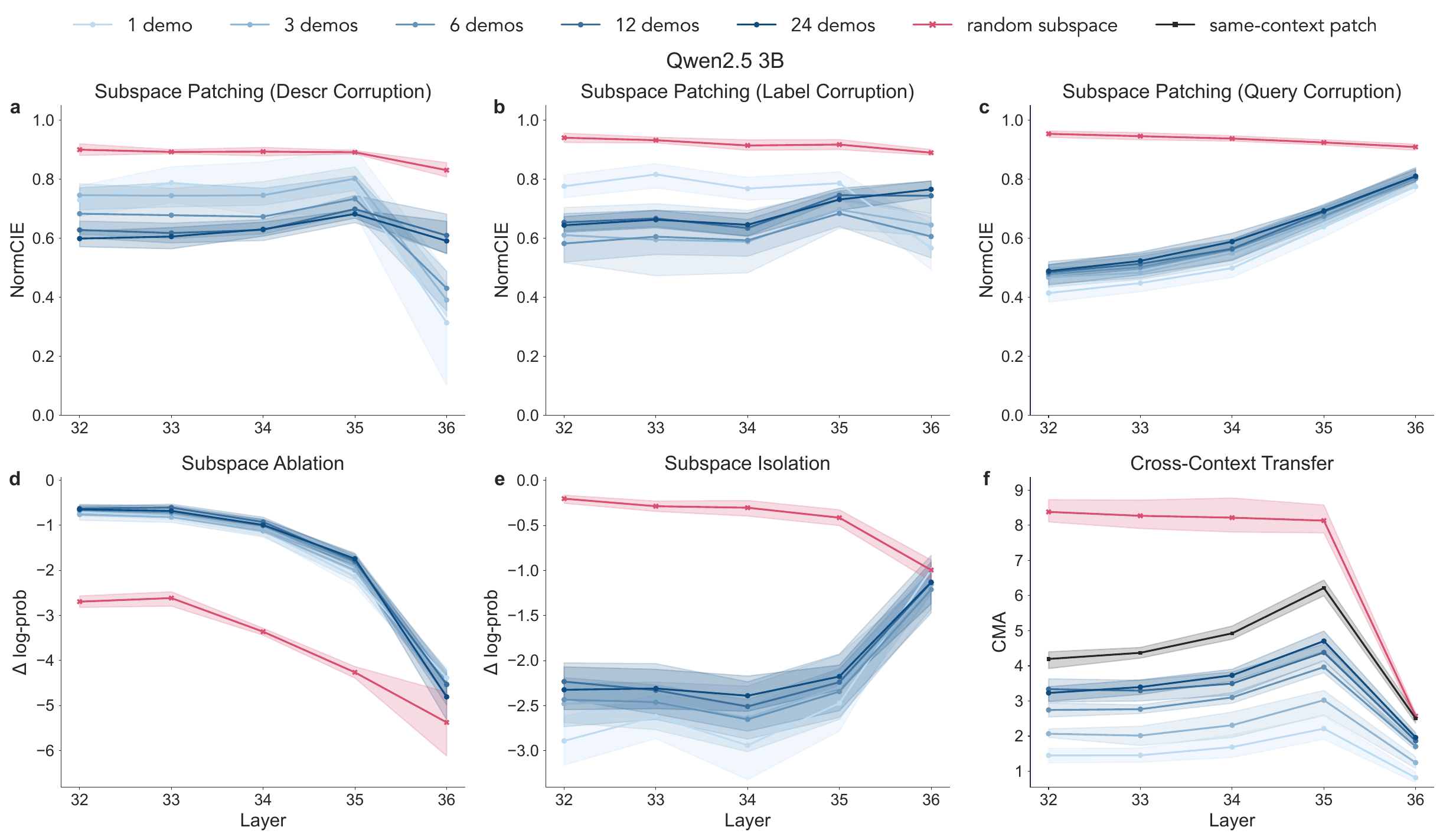}}
    \caption{Causal intervention results for Qwen2.5~3B. \textbf{a}--\textbf{c}, Activation patching with $N$ demonstrations under three corruption conditions: description (\textbf{a}), label (\textbf{b}) and query (\textbf{c}). Patching the conceptual subspace (blue) is compared against a random-subspace baseline (red). The x-axis indexes layers, and the y-axis shows the normalized causal indirect effect (CIE). \textbf{d}--\textbf{e}, Subspace necessity and sufficiency tested by ablating the conceptual subspace (\textbf{d}) or isolating it (\textbf{e}). The y-axis denotes the change in log-probability of the correct token. \textbf{f}, Causal effects of cross-context transfer, where the relational structure from a source context is adapted to a target context via an orthogonal transformation. The y-axis reports the causal mediation (CMA) score. Performance is compared against a random-subspace baseline (red) and a same-context patch reference (black) using the conceptual subspace derived from the target context itself. Shaded regions indicate 95\% CIs computed from 10,000 bootstrap resamples across five runs.}
    \label{fig:si-causality-qwen2v5_3b}
    \end{center}
\end{figure*}

Besides Llama-3.1~70B, we applied the same set of targeted causal interventions to other LLMs; the corresponding results are shown in Figures~\ref{fig:si-causality-llama3}--\ref{fig:si-causality-qwen2v5_3b}. Across models, we observed trends broadly consistent with those reported for Llama-3.1~70B, although the extent to which inference is mediated by the conceptual subspace varied across models.

Specifically, activation patching under corrupted demonstrations and queries (Figures~\ref{fig:si-causality-llama3}a--c,g--i; \ref{fig:si-causality-qwen2v5}a--c,g--i) showed that demonstration-derived contextual information was primarily mediated by the conceptual subspace at mid-depth, where the subspace began to form, whereas query-related information was mediated at relatively later layers. While patching the conceptual subspace substantially recovered performance compared to random-subspace baselines, the recovery gap was narrower than that observed in Llama-3.1~70B. This suggests that, in these smaller models, task-relevant information may be represented in a more distributed manner, with weaker reliance on a localized, dominant conceptual subspace.

Subspace ablation and isolation experiments (Figures~\ref{fig:si-causality-llama3}d--e,j--k; \ref{fig:si-causality-qwen2v5}d--e,j--k) revealed patterns similar to those in Llama-3.1~70B. Ablating the conceptual subspace consistently disrupted inference and led to a clear performance drop relative to the random-subspace baselines, although performance did not collapse entirely, indicating partial redundancy or compensatory pathways. Conversely, isolating the subspace preserved near-original performance at middle layers, with degradation in later layers where computation likely shifted toward surface-form generation. As in Llama-3.1~70B, increasing the number of demonstrations attenuated ablation effects, while isolation continued to preserve performance deeper in the model, suggesting a progressively stronger reliance on the subspace as contextual information accumulates.

Finally, cross-context transfer effects increased with the number of demonstrations and consistently exceeded the random-subspace baselines (Figures~\ref{fig:si-causality-llama3}f,l; \ref{fig:si-causality-qwen2v5}f,l), indicating that these models can construct relational structure that generalizes across contexts. However, the magnitude of this effect was notably reduced in smaller models. This trend was especially pronounced in Qwen2.5~3B (Figure~\ref{fig:si-causality-qwen2v5_3b}), where random-subspace baselines often matched or exceeded the effects of the derived conceptual subspace across interventions. Together, these results suggest a transition in representational strategy: whereas larger models leverage well-defined, abstract conceptual subspaces to approximate in-context inference, smaller models rely more heavily on diffuse or non-subspace-specific representations.

\subsection{Analyzing Context Integration into Conceptual Subspaces across Models}
\label{si:results-context}

\begin{figure*}[t]
    \begin{center}
    \centerline{\includegraphics[width=\textwidth]{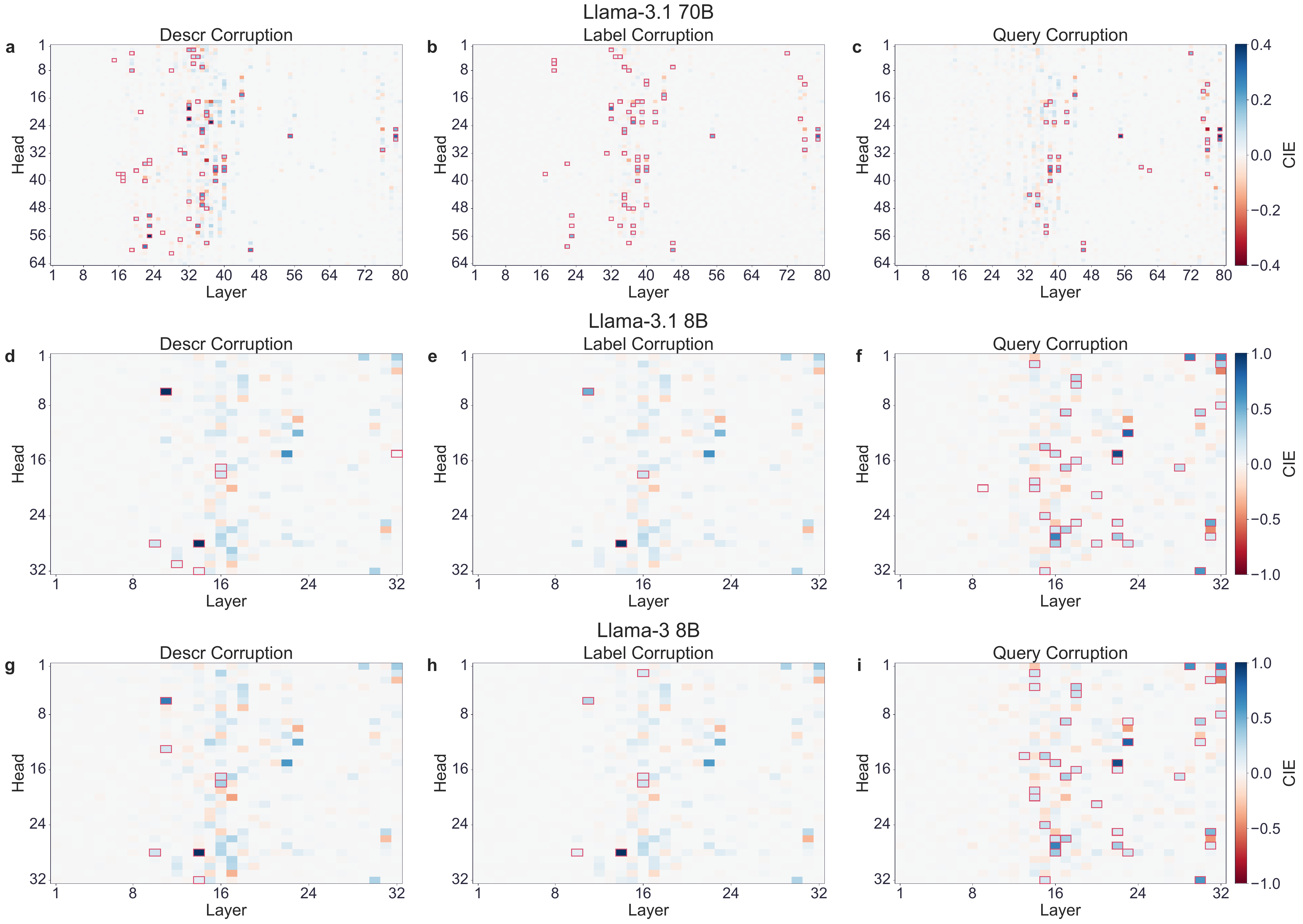}}
    \caption{Causal indirect effects (CIEs) of attention heads across Llama-3.1~70B (\textbf{a}--\textbf{c}), Llama-3.1~8B (\textbf{d}--\textbf{f}), and Llama-3~8B (\textbf{g}--\textbf{i}). Within each row, columns represent CIEs under description, label, and query corruption, respectively. The x axis indexes model layers, and the y-axis denotes attention head indices. Bordered cells highlight attention heads with statistically significant effects under the respective corruption conditions.}
    \label{fig:si-sgnf_heads-llama3}
    \end{center}
\end{figure*}

\begin{figure*}[t]
    \begin{center}
    \centerline{\includegraphics[width=\textwidth]{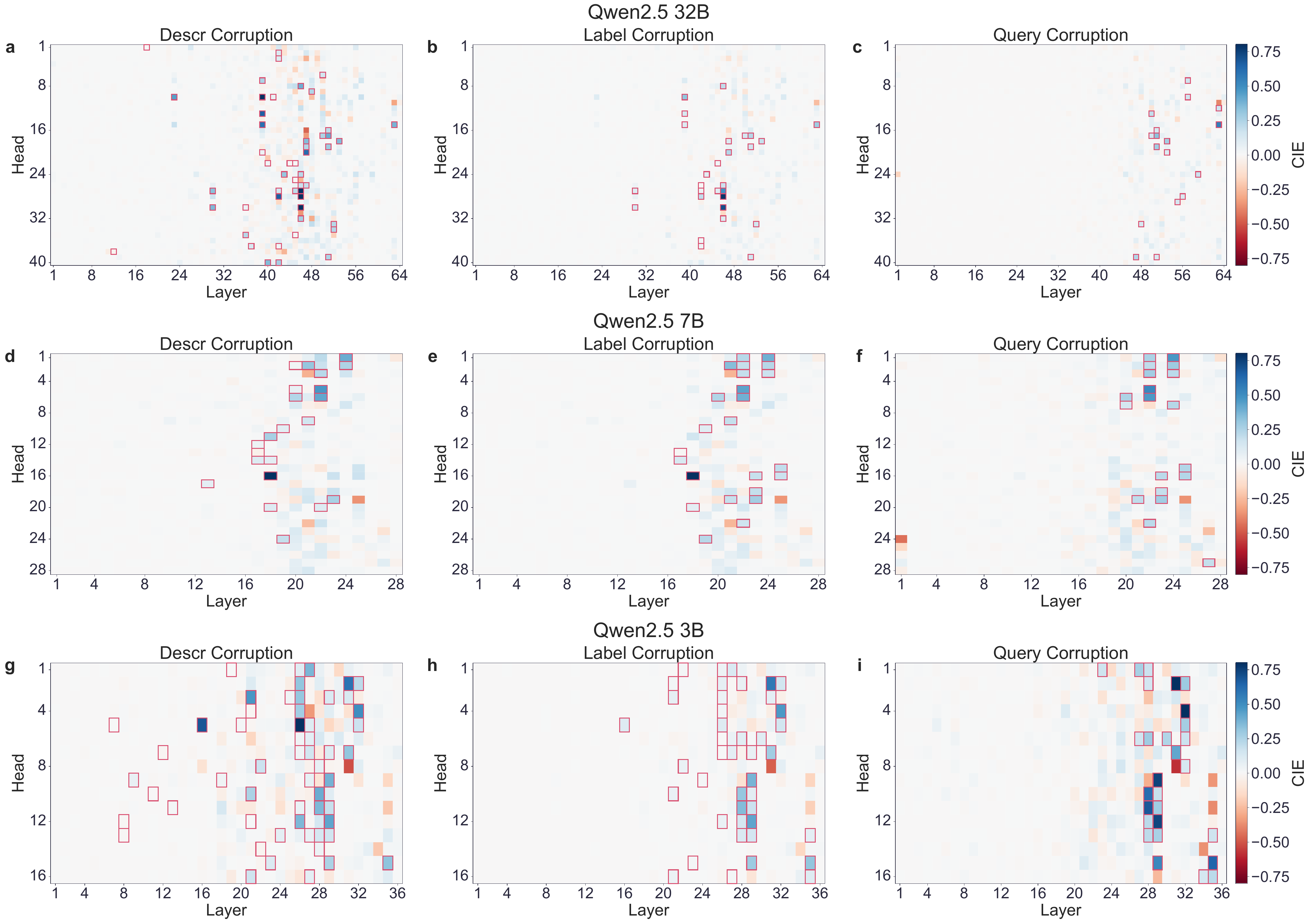}}
    \caption{Causal indirect effects (CIEs) of attention heads across Qwen2.5~32B (\textbf{a}--\textbf{c}), Qwen2.5~7B (\textbf{d}--\textbf{f}), and Qwen2.5~3B (\textbf{g}--\textbf{i}). Within each row, columns represent CIEs under description, label, and query corruption, respectively. The x axis indexes model layers, and the y-axis denotes attention head indices. Bordered cells highlight attention heads with statistically significant effects under the respective corruption conditions.}
    \label{fig:si-sgnf_heads-qwen2v5}
    \end{center}
\end{figure*}

Across the LLMs we tested, activation patching across corruption conditions show a staged progression of information integration consistent with the patterns observed in Llama-3.1~70B. Heads with statistically significant effects under demonstration corruptions (descriptions and labels) clustered in early-to-middle layers for Llama-3.1~8B and Llama-3~8B (Figure~\ref{fig:si-sgnf_heads-llama3}), yet shifted toward middle-to-late layers in the Qwen2.5 models (Figure~\ref{fig:si-sgnf_heads-qwen2v5}). These depth ranges aligned with the formation of the shared conceptual subspace and its causal mediation of contextual information (Sections~\ref{si:results-emergence}--\ref{si:results-causality}). We also observed substantial overlap between heads identified under description and label corruptions, suggesting a shared mechanism for integrating demonstration-derived cues into the conceptual subspace. In larger models (Llama-3.1~70B and Qwen2.5~32B), description corruption typically implicated more heads and yielded larger causal indirect effects (CIEs) than label corruption. This likely reflects the role of descriptions in contextualizing the query~\citep{bakalova2025contextualize}, which appears more pronounced as model capacity increases. Conversely, query corruption consistently implicated a relatively later set of heads, mirroring our observation that query information is mediated by the conceptual subspace over a later stage of processing. These mid-to-late layer heads likely support query-conditioned computations given the established context, whereas final-layer heads may be engaged in a readout stage that maps internal representations to surface-form token predictions.

\begin{figure*}[t]
    \begin{center}
    \centerline{\includegraphics[width=\textwidth]{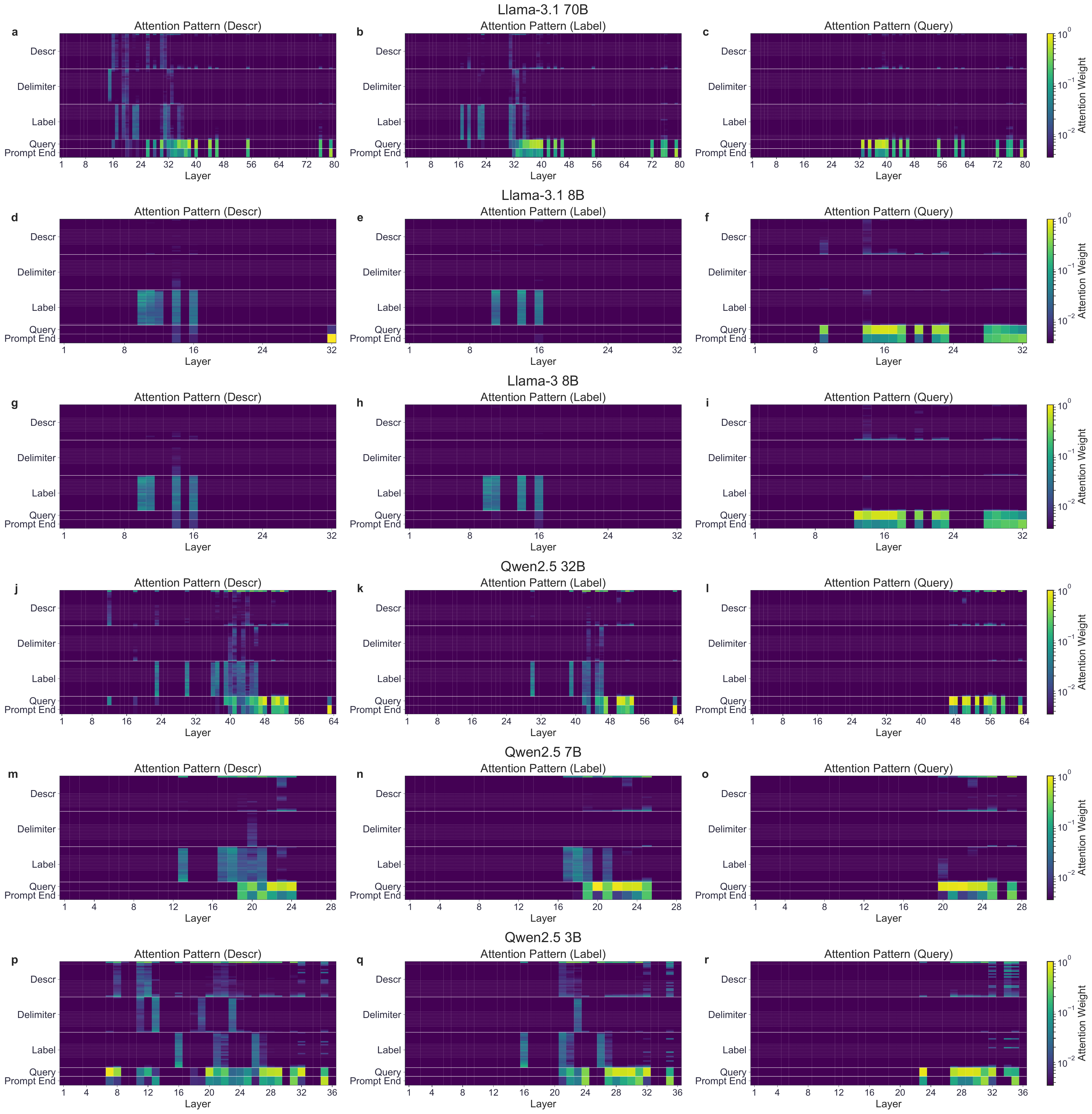}}
    \caption{Attention patterns of heads with statistically significant causal indirect effects (CIEs) in Llama-3.1~70B (\textbf{a}--\textbf{c}), Llama-3.1~8B (\textbf{d}--\textbf{f}), Llama-3~8B (\textbf{g}--\textbf{i}), Qwen2.5~32B (\textbf{j}--\textbf{l}), Qwen2.5~7B (\textbf{m}--\textbf{o}), and Qwen2.5~3B (\textbf{p}--\textbf{r}). Within each model (row), columns show the attention patterns of heads identified under description, label, and query corruption, respectively. Within each layer, attention patterns were averaged over heads with statistically significant CIEs. Layers with no significant heads were set to zero. The x-axis indicates layer index, and the y-axis shows attended token spans grouped by source segment. The description, delimiter, and label spans for the 24 in-context demonstrations are ordered top-to-bottom based on their sequence in the prompt. See Figure~\ref{fig:si-attn_pattern} for attention patterns of individual heads.}
    \label{fig:si-attn_pattern-layerwise}
    \end{center}
\end{figure*}

\begin{figure*}[t]
    \begin{center}
    \centerline{\includegraphics[width=\textwidth]{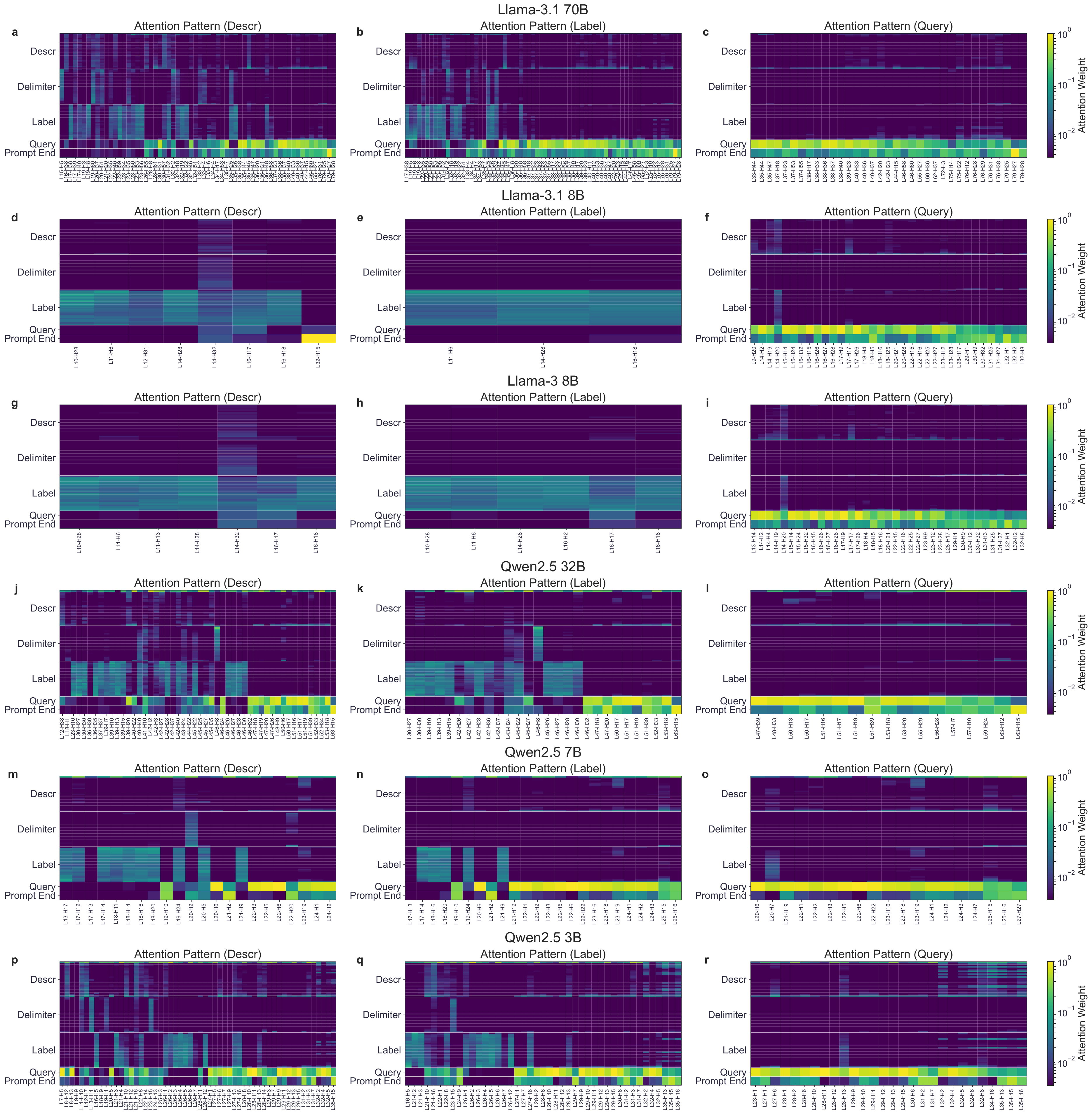}}
    \caption{Attention patterns of heads with statistically significant causal indirect effects (CIEs) in Llama-3.1~70B (\textbf{a}--\textbf{c}), Llama-3.1~8B (\textbf{d}--\textbf{f}), Llama-3~8B (\textbf{g}--\textbf{i}), Qwen2.5~32B (\textbf{j}--\textbf{l}), Qwen2.5~7B (\textbf{m}--\textbf{o}), and Qwen2.5~3B (\textbf{p}--\textbf{r}). Within each model (row), columns show the attention patterns of heads identified under description, label, and query corruption, respectively. The x-axis denotes attention heads by their (layer, head) index, and the y-axis shows attended token spans grouped by source segment. The description, delimiter, and label spans for the 24 in-context demonstrations are ordered top-to-bottom based on their sequence in the prompt.}
    \label{fig:si-attn_pattern}
    \end{center}
\end{figure*}

Attention patterns further reveal a functional shift from demonstration-centric to query-centric processing (Figures~\ref{fig:si-attn_pattern-layerwise}--\ref{fig:si-attn_pattern}). Heads identified under demonstration corruptions and located in relatively early layers allocated most attention to descriptions, delimiters, and labels, consistent with extracting and integrating demonstration cues to construct or update the subspace. In contrast, later-layer heads shifted attention toward the query and the final prompt delimiter, signaling a transition from subspace construction to query-conditioned inference. Heads identified under query corruption attended heavily to query tokens, while final-layer heads attended more strongly to the last delimiter, consistent with next-token prediction.

\begin{figure*}[t]
    \begin{center}
    \centerline{\includegraphics[width=0.9\textwidth]{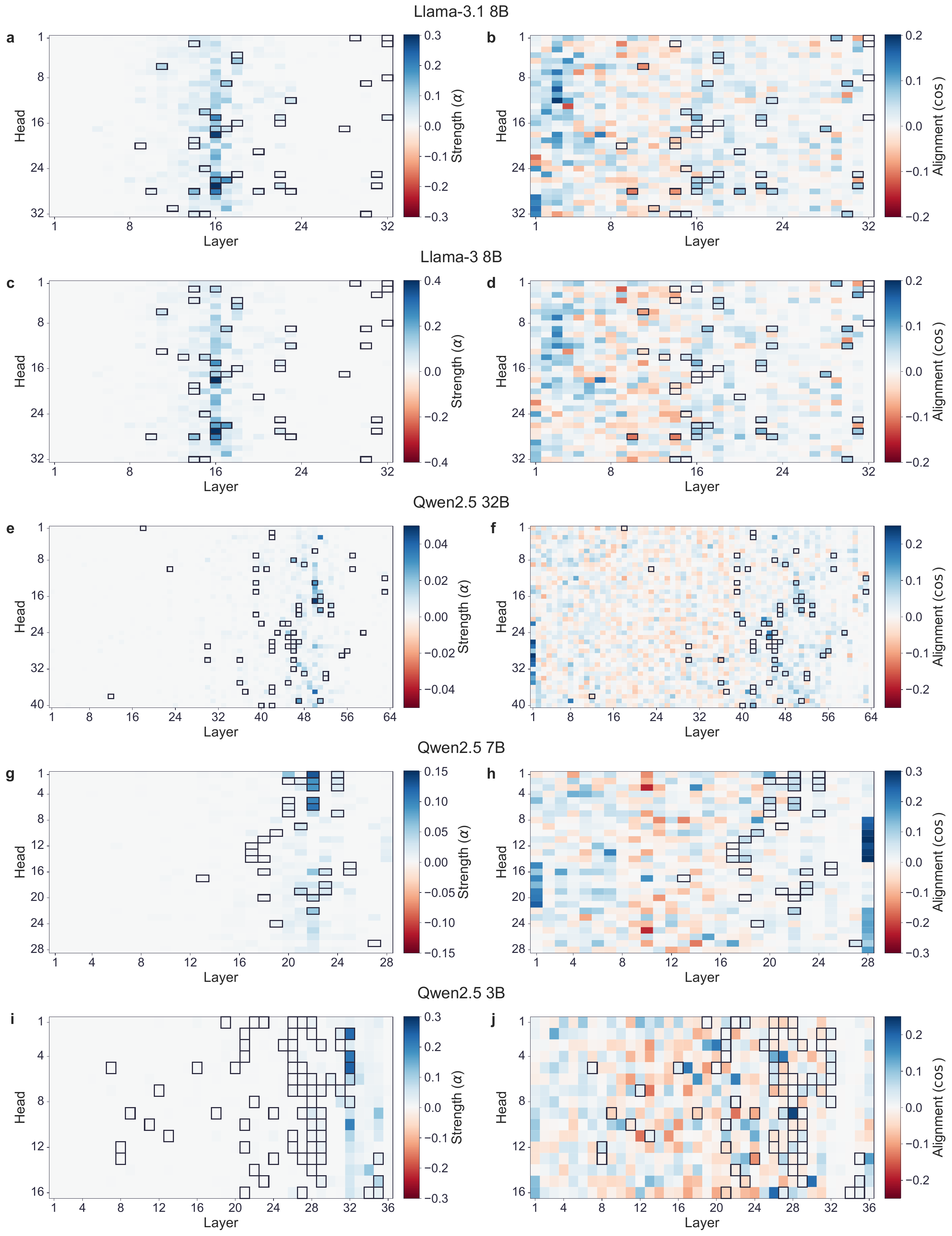}}
    \caption{Contribution of attention heads to the conceptual subspace across Llama-3.1~8B (\textbf{a}--\textbf{b}), Llama-3~8B (\textbf{c}--\textbf{d}), Qwen2.5~32B (\textbf{e}--\textbf{f}), Qwen2.5~7B (\textbf{g}--\textbf{h}), and Qwen2.5~3B (\textbf{i}--\textbf{j}). For each model, the left column reports the contribution strength ($\alpha$) of each head to the conceptual subspace, and the right column shows the directional alignment (cosine similarity) between head outputs and the subspace. Bordered cells highlight attention heads with statistically significant causal indirect effects (CIEs) under any corruption condition.}
    \label{fig:si-attn_alpha_align}
    \end{center}
\end{figure*}

\begin{figure*}[t]
    \begin{center}
    \centerline{\includegraphics[width=\textwidth]{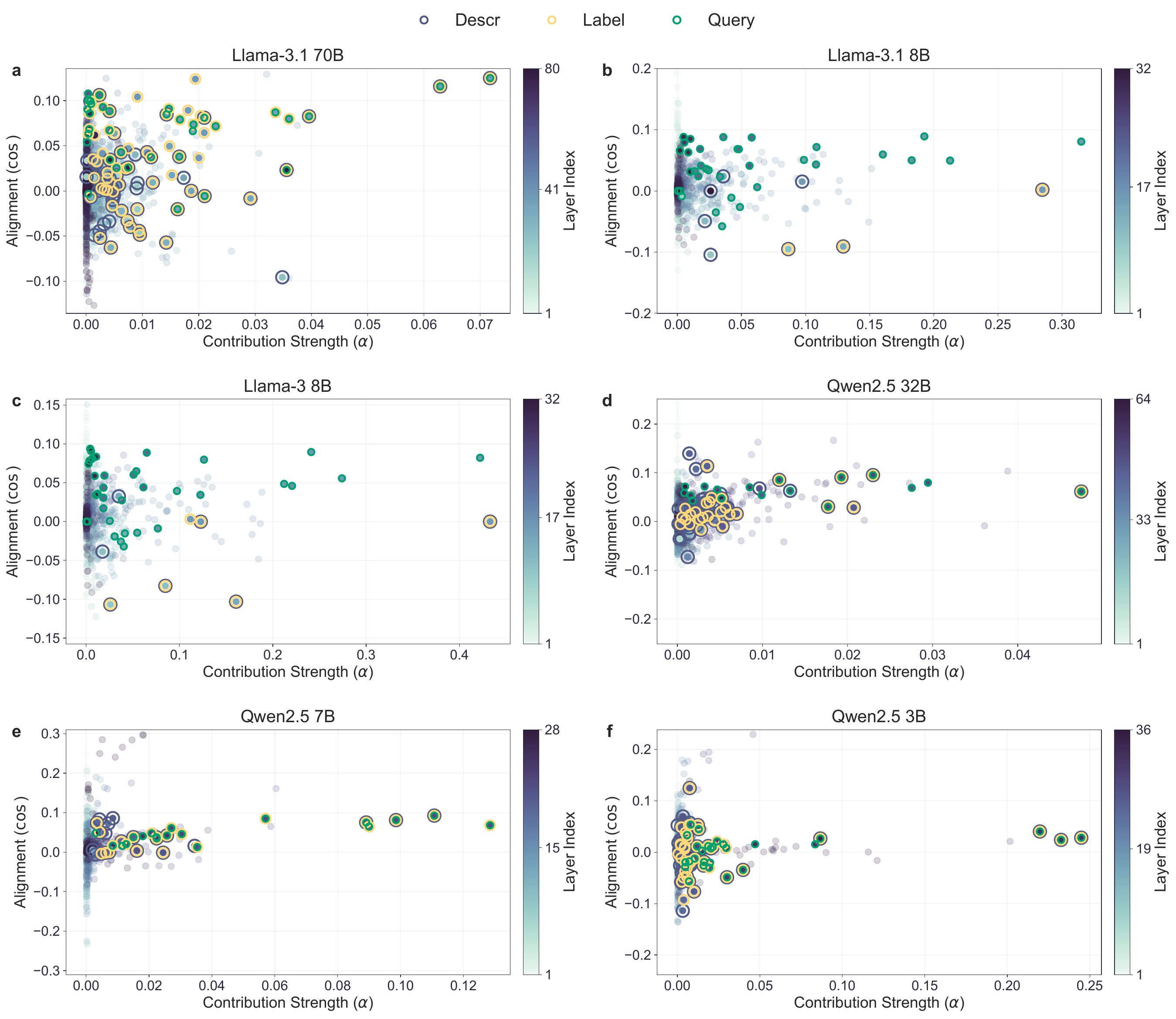}}
    \caption{Contribution--alignment relationship for attention heads across Llama-3.1 70B (\textbf{a}), Llama-3.1 8B (\textbf{b}), Llama-3 8B (\textbf{c}), Qwen2.5 32B (\textbf{d}), Qwen2.5 7B (\textbf{e}), and Qwen2.5 3B (\textbf{f}). Each point corresponds to an attention head, with the x-axis denoting its contribution strength to the conceptual subspace ($\alpha$) and the y-axis showing its directional alignment (cosine similarity) with that subspace. Points are colored by layer index. Outlines mark heads with statistically significant causal indirect effects (CIEs); the outline color indicates the corruption condition(s) (description, label, query) under which significance was detected.}
    \label{fig:si-attn_alpha_vs_align}
    \end{center}
\end{figure*}

The identified heads were also geometrically aligned with the conceptual subspace (Figures~\ref{fig:si-attn_alpha_align}--\ref{fig:si-attn_alpha_vs_align}). Significant heads in early-to-middle layers of Llama-3 (or middle-to-late layers of Qwen2.5) exhibited higher contribution strength ($\alpha$) and stronger directional alignment ($\textrm{align}$) than non-significant heads. While contribution strength generally decreased in later layers, directional alignment remained relatively high, suggesting that late-layer computations stay coordinated with an already-established subspace (Figure~\ref{fig:si-attn_alpha_align}). Figure~\ref{fig:si-attn_alpha_vs_align} further shows a qualitative difference in how heads interact with this subspace: heads identified under query corruption predominantly wrote coherently into the conceptual subspace (positive directional alignment), suggesting a role in using or stabilizing the subspace for prediction. In contrast, heads identified under demonstration corruption can exhibit substantial write strength ($\alpha$) but weaker alignment, indicating a role in constructing or reconfiguring the subspace based on contextual evidence rather than simply amplifying existing directions. 

In larger models (Llama-3.1~70B and Qwen2.5~32B), we observed notable overlap among the heads identified across corruption conditions; these heads exhibited both high contribution strength and high alignment (Figure~\ref{fig:si-attn_alpha_vs_align}a,d), indicating a core set of components that integrate relevant contextual information and directly drive model inference. In smaller models, causally relevant heads were more diffuse and can show weaker alignment or contribution strength relative to other heads (Figures~\ref{fig:si-attn_alpha_vs_align}b--c,e--f; \ref{fig:si-attn_alpha_align}g--i). Finally, individual heads in the larger models tended to have lower contribution strengths than those in smaller models, suggesting that computation is distributed across a broader ensemble of moderately weighted heads.

Taken together, these findings support a two-stage account that generalizes across model families: (i) Construction, in which earlier layers integrate demonstration-derived cues to build and stabilize a conceptual subspace; and (ii) Utilization, in which later layers operate within this structured subspace to perform query-conditioned inference and generate the final prediction. Smaller models appear to rely less consistently on this mechanism, exhibiting more diffuse, non-subspace-specific strategies.

\begin{figure}[t]
\begin{center}
\centerline{\includegraphics[width=0.55\linewidth]{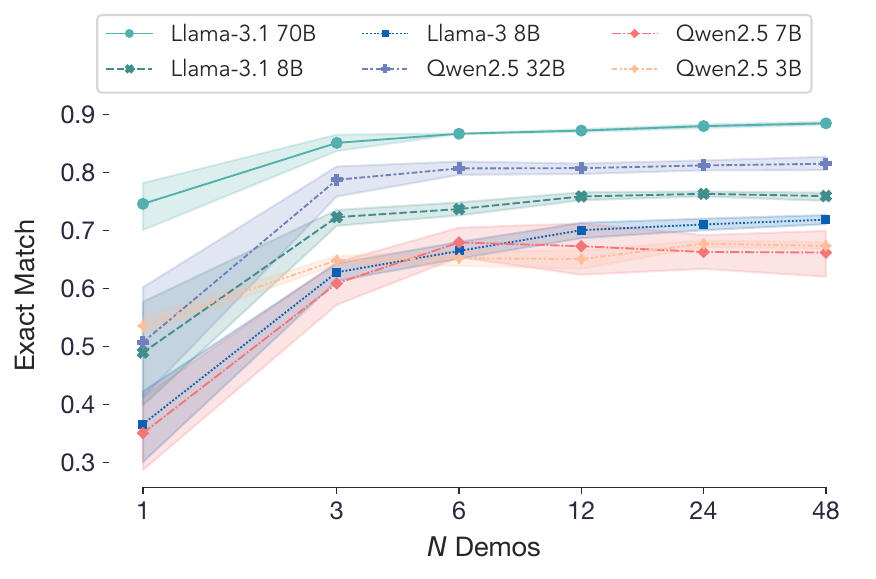}}
\caption{Performance of various LLMs on the reverse dictionary task, evaluated on the THINGS data and measured through exact match accuracy. The models were presented with $N$ demonstrations sampled from the training set and evaluated on an independent test set. Shaded areas denote $95\%$ confidence intervals, calculated from 10,000 resamples across five independent runs.}\label{fig:si-behavior}
\end{center}
\end{figure}

\end{document}